\documentclass{article}

\usepackage{lipsum}

\usepackage{microtype}
\usepackage{graphicx}
\usepackage{subcaption}
\usepackage{booktabs} % for professional tables
\usepackage{hyperref}

\usepackage[accepted]{icml2026}

\usepackage{amsmath}
\usepackage{amssymb}
\usepackage{mathtools}
\usepackage{amsthm}
\usepackage[capitalize,noabbrev]{cleveref}

\theoremstyle{plain}
\newtheorem{theorem}{Theorem}[section]
\newtheorem{proposition}[theorem]{Proposition}

\theoremstyle{definition}

\theoremstyle{remark}

\usepackage[textsize=tiny]{todonotes}

\usepackage{bbm}
\usepackage{float}
\usepackage{tikz}
\usepackage[dvipsnames, table]{xcolor}
\usetikzlibrary{arrows}        
\usepackage{subcaption} 
\usepackage{multirow}
\usepackage[T1]{fontenc}

\newcommand\methodname{DisCoVR}
\DeclareMathOperator{\ELBO}{ELBO}
\newcommand\KL{\mathcal{D}_{\mathrm{KL}}}
\newcommand\E{\mathbb{E}}

\definecolor{LowerBase}{RGB}{230, 69, 46}
\definecolor{UpperBase}{RGB}{27, 154, 139}
\colorlet{Lower}{LowerBase!55}
\colorlet{Upper}{UpperBase!55}

\icmltitlerunning{Variational Learning of Disentangled Representations}

\begin{document}

\twocolumn[
  \icmltitle{Variational Learning of Disentangled Representations}
  \icmlsetsymbol{equal}{*}

  \begin{icmlauthorlist}
    \icmlauthor{Yuli Slavutsky}{equal,stats}
    \icmlauthor{Ozgur Beker}{equal,irv,cs}
    \icmlauthor{David M. Blei}{stats,cs}
    \icmlauthor{Bianca Dumitrascu}{irv,stats,cs,cell}
  \end{icmlauthorlist}

  \icmlaffiliation{stats}{Department of Statistics, Columbia University, New York City, USA}
  \icmlaffiliation{irv}{Irving Institute for Cancer Dynamics, Columbia University, New York City, USA}
  \icmlaffiliation{cs}{Department of Computer Science, Columbia University, New York City, USA}
  \icmlaffiliation{cell}{Columbia Stem Cell Initiative, Columbia University, New York City, USA}

  \icmlcorrespondingauthor{Bianca Dumitrascu}{bmd2151@columbia.edu}
  \icmlcorrespondingauthor{Yuli Slavutsky}{ys3938@columbia.edu}
  \icmlkeywords{Machine Learning, ICML}
  \vskip 0.3in
]

\printAffiliationsAndNotice{}

\begin{abstract}
  Disentangled representations separate factors that are shared across conditions from those that are condition-specific. Such separation is needed for generalization to new domains, treatments, patients, or species. A dominant line of work pursues this goal through variational formulations. While these approaches achieve partial disentanglement, they often exhibit three common limitations: they either do not remove all condition-specific information from the condition-specific representation, allow the condition-specific representation to become uninformative, or impose independence assumptions that do not reflect the underlying generative process.
  In this work, we introduce DisCoVR, a variational framework that addresses these limitations. Its objective is aligned with the probabilistic structure of the data-generating process, and includes an adversarial term that prevents condition-specific information from being encoded in the condition-specific representation.
  DisCoVR reconstructs the data from both shared and condition-specific representations, ensuring that each remains informative, and uses a structured prior that further reinforces the informativeness of both representations.
  We show that across synthetic, image, and single-cell RNA-sequencing datasets, DisCoVR achieves stronger disentanglement compared to previous approaches.
\end{abstract}

\section{Introduction} \label{intro}

Neural network–based models excel at learning rich representations of complex data, and are increasingly used in settings where each observation $x \in \mathcal{X} \subseteq \mathbb{R}^d$ is paired with a condition label $y \in \{1, \dots, K\}$, such as   patient, site, or experimental condition.
Generalizing these representations to new domains often requires disentangling factors shared across conditions from those specific to each $y$.

Generative models provide a natural framework for uncovering latent structure and learning data representations, with prominent examples including Generative Adversarial Networks (GANs) \citep{goodfellow2020generative}, Variational Autoencoders (VAEs) \citep{vae}, and diffusion models \citep{sohl2015deep, ho2020denoising}. Among these, VAEs and their extensions are particularly well-suited to transfer learning and domain adaptation \citep{akrami2020brain, lovric2021should, godinez2022design, zhang2023variational}, thanks to their probabilistic foundation and ability to capture uncertainty.

Accordingly, several VAE-based methods have been proposed to integrate data across multiple conditions or sources \citep{xu2021probabilistic, lotfollahi2019scgen, boyeau2022deep}, but only a few explicitly disentangle invariant and condition-specific components \citep{CVAE, CSVAE, CCVAE, diva}. 

While these approaches improve separation to some degree, they either (i) leave the shared component free to retain label information without an explicit mechanism enforcing invariance \citep{CVAE,diva}, (ii) reconstruct $x$ jointly from the invariant and condition-specific components, so that the invariant one can remain uninformative \citep{CSVAE}, (iii) impose independence assumptions that misalign with the underlying generative structure \citep{diva}, or (iv)  encode in their priors the assumption that the invariant representation in fact can vary with the label \citep{CCVAE}.

In this work, we introduce a framework for learning \emph{disentangled representations in multi-condition datasets} that addresses these limitations: (a) We formulate a principled probabilistic objective that encodes the correct conditional independencies. (b) We specify a prior structure in which the condition-specific representation $w$ depends on the mean of the invariant representation $z$. Since $w$ depends only on class-specific aggregation of $z$, the invariant representation cannot absorb condition-specific leakage from $w$, while forcing $z$ to remain informative. (c) Our method uses reconstruction paths from both representations, which further enforces informativeness of $z$.
(d) The resulting optimization objective includes an adversarial term that explicitly discourages leakage of condition-specific information into the invariant representation. 

Our approach formulates disentanglement as a max–min game, which we show to admit a unique equilibrium. Empirically, we show through extensive experiments on synthetic benchmarks and real-world datasets, that our method consistently improves over existing approaches in disentangling shared and condition-specific structure.

\section{DisCoVR: Disentangling Common and Variant Representations} \label{model}

For the task of learning disentangled representations from multi-condition data,
we consider a dataset $\mathcal{D} = \{(x_i, y_i)\}_{i=1}^n$ consisting of inputs $x_i \in \mathcal{X} \subseteq \mathbb{R}^d$ collected from associated condition labels $y_i \in \{1, \dots, K\}$. For each class $k$ (corresponding to a study or experimental condition), the associated subset $\mathcal{D}_k \coloneqq \{x_i: y_i=k\}$ consists of i.i.d.\ samples drawn from a class-conditional distribution $p(x\mid y=k)$. 

\subsection{Model assumptions} \label{sec:assump}

We assume that the data is generated by latent variables $z$ and $w$, such that 
 the joint distribution $p(x,y,z,w)$  factorizes according to the probabilistic graphical model illustrated in Figure \ref{fig:prob_graph}, i.e.,
\begin{equation}
    p(x,y,z,w) = p(y) \, p(w \mid y) \, p(z) \, p(x \mid z, w).
\end{equation}

This model encodes two key conditional independence assumptions:
\begin{enumerate}
    \item \emph{Latent variable conditional independence:} 
     Given the condition $y$, the latent representations $z$ and $w$ are conditionally independent: $z\perp w\mid y$. \label{assump1}
    \item \emph{Sufficiency of the condition-aware latent representation:} The input $x$ is conditionally independent of the condition $y$ given $w$: $x\perp y \mid w$. \label{assump2}
\end{enumerate}

Note that in this formulation, $z$ and $w$ are no longer independent if conditioned also on $x$, that is, $z \not \perp w \mid x, y$. 

\begin{figure}[ht]
\centering
\begin{tikzpicture}

\tikzset{
circ/.style={circle,  draw=black, minimum size=6mm, inner sep=0pt},
sqr/.style={rectangle, draw=black, minimum size=6mm, inner sep=0pt}
}

% --- nodes
\node[circ, fill=gray!20] (y) at (0,0)        {$y$};
\node[circ]  (w) at (1.5,0)      {$w$};
\node[circ]  (z) at (1.5,-1)     {$z$};
\node[circ, fill=gray!20] (x) at (3.0,-0.5)   {$x$};

% --- edges (all H/V)
\draw[->] (y) -- (w);
\draw[->] (w) -| (x);
\draw[->] (z) -| (x);

\end{tikzpicture}
\caption{Probabilistic graphical model: gray circles denote observed variables, white circles show latent variables.}
\label{fig:prob_graph}
\end{figure}

\subsection{Target posterior structure} 

In our model, each observation $x$ is generated from two latent variables: $z$, which is \emph{condition-invariant}, and $w$, which is \emph{condition-aware} through its dependence on $y$. Our goal is to learn probabilistic representations where the marginals of $z$ and $w$ preserve this structure, yielding disentangled factors. That is, we aim to approximate the posterior $p_{z,w \mid x,y}$.

However, approximating the full posterior with a variational distribution $q_{z, w \vert x, y}$ is intractable: even for simple variational families such as Gaussians, modeling the dependencies between $z$ and $w$ requires a full covariance structure, which is computationally prohibitive in high dimensions.
To mitigate this, we employ a factorized approximation $q_{z\vert x} \ q_{w \vert x,y}$.

Our variational approximation is guided by two complementary objectives: (i) $q_{z\vert x}$  closely approximating the marginal posterior $p_{z\vert x}$; and (ii) the product $q_{z\vert x} \ q_{w \vert x,y}$ closely approximating the true posterior $p_{z, w \vert x, y}$. 
Formally, given variational families\footnote{Here we consider general families and specify our concrete choices in \S \ref{sec:priors}.} $\mathcal{Q}_z$ and $\mathcal{Q}_w$ we seek to find $q^*_{z\vert x} \in \mathcal{Q}_z$ and $q^*_{w \vert x,y} \in \mathcal{Q}_w$ that minimize the following sum of Kullback-Leibler (KL) divergences: 
\begin{align}\label{eq:KLs}
q_{z\vert x}^{*},\ q_{w\vert x,y}^{*}=\arg\min_{\substack{q_{z\vert x}\in\mathcal{Q}_{z}\\
q_{w\vert x,y}\in\mathcal{Q}_{w}
}
} & \left[\KL(q_{z\vert x}\ \Vert\ p_{z\vert x})\right.\\ \notag
 & \left.+\KL(q_{z\vert x}\,q_{w\vert x,y}\ \Vert\ p_{z,w\vert x,y})\right].
\end{align}

\subsection{Optimization objective} 

Since direct evaluation of the KL divergences in Equation~\ref{eq:KLs} is intractable, we optimize a surrogate objective consisting of two ELBO terms.

The corresponding ELBO objective to minimize $\KL(q_{z\vert x} \ \Vert \ p_{z\vert x})$ is
\begin{equation} \label{eq:elbo_z}
    \mathcal{L}_z(q_{z\vert x}, p; x) \coloneqq \E_{q_{z \vert x}}\left[\log p(x \mid z)\right] - \KL(q_{z\vert x} \, \Vert \, p_z),
\end{equation}
and the ELBO objective for the second KL term, $\KL(q_{z\vert x}q_{w \vert x,y} \ \Vert \ p_{z, w \vert x, y})$, is
% \hspace{-1em}
% \begin{align} \label{eq:elbo_w}
% \textstyle
% \mathcal{L}_w (q_{w\vert x,y},p;x,y)\! \coloneqq & \E_{q_{z\vert x}} \! \left[\E_{q_{w\vert x,y}} \! \left[\log\left(p(x\vert z,w)\right)\right]\right] \!\! -\! \KL\left(q_{z\vert x} \, \Vert  \, p_z\right)
% \!\!- \! \KL\left(q_{w \vert x,y} \, \Vert \, p_{w\mid y}\right).
% \end{align}
\begin{align} \label{eq:elbo_w} \notag
 & \mathcal{L}_{w}(q_{w\vert x,y},p;x,y)\!\coloneqq\E_{q_{z\vert x}}\!\left[\E_{q_{w\vert x,y}}\!\left[\log\left(p(x\vert z,w)\right)\right]\right]\!\!\\ 
 & -\!\KL\left(q_{z\vert x}\,\Vert\,p_{z}\right)\!\!-\!\KL\left(q_{w\vert x,y}\,\Vert\,p_{w\mid y}\right).
\end{align}

Note that  $\mathcal{L}_w(q_{w\vert x,y},p;x,y)$ is the ELBO objective that corresponds to a factorized posterior $q_{z\vert x}q_{w \vert x,y}$. In  Proposition \ref{prop:obj_gap} we examine the gap between this objective and an ELBO  corresponding to a full variational posterior. This can be interpreted as the cost of enforcing a condition-invariant latent representation, specifically, constraining $z$ to depend only on $x$.
\begin{proposition} \label{prop:obj_gap}
    For random variables $x, y, z, w$ following the graphical model in Figure \ref{fig:prob_graph},
    \begin{align*}
     & \ELBO(q,p;x,y)-\mathcal{L}_{w}(q_{w\vert x,y},p;x,y)\\
     & =\E_{q_{w\vert x,y}}\left[\text{KL}\left(q_{z\vert x}\,\Vert\,p_{z\vert w,x,y}\right)\right]
    \end{align*}
    where 
    \begin{equation*}
        \ELBO(q,p;x,y) \coloneqq \log p(x\mid y)-\KL\left(q_{w\vert x,y}\,\Vert\,p_{w\vert x,y}\right).
    \end{equation*}
\end{proposition}
The proof is provided in Appendix \ref{sup:obj_gap}.

Note that a full definition of the objectives in Equations \ref{eq:elbo_z} and \ref{eq:elbo_w} requires the specification of corresponding prior distributions, namely $p_z$ and $p_{w \vert y}$. We defer their definitions to \S \ref{sec:priors}. 

Equation \ref{eq:elbo_w} provides an evidence lower bound on the conditional log-likelihood $\log p(x \mid y)$. By adding $\log p(y)$, this bound extends to the joint log-marginal likelihood $\log p(x, y)$. Beyond optimizing this objective, we  aim to ensure that the marginal distribution over $y$ implicitly induced by the latent representations is consistent with the true $p(y)$.

To this end, we introduce an auxiliary classifier $g(y \mid z)$ as a form of posterior regularization \citep{ganchev2010posterior}. This classifier captures the residual predictive signal about $y$ in $z$ and is trained by minimizing the expected negative log-likelihood $-\E_{q(z \mid x)} \log g(y \mid z)$. If $z$ is truly independent of $y$, then $g(y \mid z)$ will approximate the marginal distribution $p(y)$. By penalizing deviations from this behavior, we enforce the structural constraint $z \perp y$ in the learned representation.

For this term to effectively encourage $q_{z\vert x}$ to discard condition-specific information, the classifier $g_{y \vert z} \in \mathcal{G}$ must be trained adversarially, with its own update step. This prevents degenerate solutions in which the loss is minimized without  removing information about $y$ from $z$, for example, by collapsing $g$ to a constant predictor that ignores its input.

Combining the three terms above, we define the objective
\begin{align}
 % \notag
 & \mathcal{L}(q_{z\vert x},q_{w\vert x,y},g_{y\vert z};x,y)\\ \notag
 & =\mathcal{L}_{z}(q_{z\vert x},p;x)+\mathcal{L}_{w}(q_{w\vert x,y},p;x,y)-\E_{q_{z\vert x}}\log g(y\mid z),
\end{align}
which can be explicitly expressed as 
\begin{align}
 & \mathcal{L}(q_{z\vert x},q_{w\vert x,y},g_{y\vert z};x,y)\coloneqq\E_{q_{z\vert x}}[\log p(x\mid z)]\\ \notag
 & \phantom{=}+\E_{q_{z\vert x}}\left[\E_{q_{w\vert x,y}}[\log p(x\mid z,w)]\right]-\E_{q_{z\vert x}}[\log g(y\mid z)]\\ \notag
 & \phantom{=}-2\KL(q_{z\vert x}\,\Vert\,p_{z})-\KL\left(q_{w\vert x,y}\,\Vert\,p_{w\mid y}\right).
\end{align}
Finally, to enable flexible trade-offs between reconstruction expressiveness and disentanglement,  we introduce weighting terms $\alpha = (\alpha_1, \alpha_2)$ into the objective following the motivation of $\beta$-VAEs \citep{beta_vae}:
\begin{align}
 & \mathcal{L}_{\alpha}(q_{z\vert x},q_{w\vert x,y},g_{y\vert z};x,y)\coloneqq\E_{q_{z\vert x}}[\log p(x\mid z)]\\ \notag
 & \phantom{=}+\E_{q_{z\vert x}}\left[\E_{q_{w\vert x,y}}[\log p(x\mid z,w)]\right]-\E_{q_{z\vert x}}\log g(y\mid z)\\ \notag
 & \phantom{=}-\alpha_{1}\KL(q_{z\vert x}\,\Vert\,p_{z})-\alpha_{2}\KL\left(q_{w\vert x,y}\,\Vert\,p_{w\mid y}\right).
\end{align}
Accordingly, the mean weighted objective is suitable for  max-min optimization of the form:
\vspace{-1.5em}
\begin{align} 
\tag{\theequation}\refstepcounter{equation}\label{eq:maxmin} \\ \notag
    \max_{q_{z\vert x}\in\mathcal{Q}_{z}}
    \max_{q_{w\vert x,y}\in\mathcal{Q}_{w}}
    \min_{g_{y\vert z}\in\mathcal{G}} 
    \E_{p_{x, y}}  \! \left[\mathcal{L}_\alpha(q_{z\vert x},q_{w\vert x,y},g_{y\vert z}; x,y) \right]. 
\end{align}

\subsection{Latent prior models and variational approximations} \label{sec:priors} 

\paragraph{Prior specification}
We place a standard Normal prior over the condition-invariant latent variable, $p_z=\mathcal{N}(0,I)$,  which reflects a non-informative prior belief over its values.

For the condition-aware latent variable $w$, we define a class-conditional Gaussian prior $p_{w\vert y}$. As a simple choice, we let $w$ have the same dimensionality as $z$ and specify
\begin{equation}
p(w\mid y=k)\!=\!\mathcal{N}(\mu_{k},I),\quad \mu_{k}\! \coloneqq \!
\E_{p_{x\vert y=k}} \! \left[\E_{q_{z\vert x}}[z]\right].
\end{equation}
Here $\mu_k$ is the mean of the inferred latent representations $z$ within the $k$-th class\footnote{Similarly, if $z$ and $w$ have different dimensions, the mean aggregation can be replaced with a neural network that maps the inferred representations $z$ for each class to the parameters of the Gaussian prior.}. 

This specification induces a coupling between the two latent variables through the data distribution.
Aligning $p_{w\vert y}$ with the class-wise expectations of the invariant variable, further encourages $q_{z\vert x}$ to encode informative representations, since capturing the shared structure will now not only increase $\mathcal{L}_z(q_{z\vert x},p;x)$, but also decrease 
$\KL\left(q_{w\mid x,y} \,\Vert\, p_{w\vert y}\right)$, and as a result increase $\mathcal{L}_w(q_{w\vert x,y},p;x,y)$.

However, this coupling between $z$ and $w$ is not defined at the level of individual samples: The prior for $w$ depends on $z$ only through the class-wise mean $\mu_y = \mathbb{E}_{p(x \mid y)} \mathbb{E}_{q(z \mid x)}[z]$, 
not through a given $z$.
Similarly, $q_{z \mid x}$ never conditions on $w$ or $y$, so the invariant representation cannot absorb condition-specific leakage from $w$.

Importantly, for a truly condition-agnostic $q_{z\vert x}$, the conditional expectations $\mu_k$ will collapse to a shared mean $\mu \coloneqq \E_{p_x}\left[\E_{q_{z \vert x}}\left[z\right]\right]$. In this case $p_{w\vert y}$ becomes a shared prior across classes, centered at a meaningful point in the latent space, rather than an uninformative one.

As the following proposition establishes, this anchoring of the prior $p_{w \vert y}$ in the variational distribution $q_{z\vert x}$ 
preserves the convex–concave structure of the objective, ensuring that the resulting max-min problem has a unique optimal solution.
\begin{proposition}
 \label{prop:optimal}
  Let $\mathcal{Q}_z$ and  $\mathcal{Q}_w$
  be convex parametric families of variational distributions over $z$ and $w$, respectively, and let $\mathcal{G}$ denote a convex set of classifiers such that $g(x) \in [0,1]$ for all $g \in \mathcal{G}$.
  Assume the latent priors are given by $z \sim p(z)$ and $p(w\vert y) =  \mathcal{N}(\mu_y,I)$, where $p(z)$ is a continuous strictly positive distribution, and $\mu_{y}=\E_{p_{x\vert y}}\left[\E_{q_{z\vert x}}[z]\right]$.
  Then, under standard regularity conditions (see Appendix \ref{sup:reg}), there exists a unique saddle point:
    \begin{align*}
     & \left(q_{z\vert x}^{*},q_{w\vert x,y}^{*},g_{y\mid z}^{*}\right)\\
     & =\max_{q_{z\vert x}\in\mathcal{Q}_{z}}\max_{q_{w\vert x,y}\in\mathcal{Q}_{w}}\min_{g_{y\vert z}\in\mathcal{G}}\mathcal{L}(q_{z\vert x},q_{w\vert x,y},g_{y\vert z}).
\end{align*}
\end{proposition}
The proof is provided in Appendix \ref{sup:opt}.

\paragraph{Specification of variational families} We set both variational families $\mathcal{Q}_z$ and $\mathcal{Q}_w$ as  
$d$-dimensional Gaussian distributions with diagonal covariance matrices. Accordingly, each variational distribution is parameterized by a mean vector $\mu \in \mathbb{R}^d$ and a vector of variances $\sigma^2 \in \mathbb{R}^d_{+}$ corresponding to the diagonal of the covariance matrix, yielding $\theta = (\mu, \sigma^2)$.

\section{Encoder-decoder model} \label{encdec}

\begin{figure}[ht]
\centering
\begin{tikzpicture}

\tikzset{
  circ/.style={circle, draw=black, fill=gray!20,
               minimum size=6mm, inner sep=0pt},
  sqr/.style={rectangle, draw=black,
              minimum height=6mm, inner sep=1pt},
  wide/.style={minimum width=6mm}
}

% -------- nodes ------------------------------------------------
\node[circ]                                                               (y)      at (0,   0) {$y$};
\node[sqr, wide]                                                          (wxy)    at (1.5, 0) {$\,w\mid x,y\,$};
\node[sqr, wide]                                                          (zx)     at (3,   0) {$\,z\mid x\,$};
\node[sqr, wide, fill=NavyBlue!20]                                        (hatx)   at (4.5, 0) {$\,\hat{x}\mid z\,$};
\node[sqr, wide, draw=NavyBlue!60!black, line width=0.8pt]                (haty)   at (6,   0) {$\hat{y}$};
\node[circ]                                                               (x)      at (3,   1) {$x$};
\node[sqr, wide, fill=NavyBlue!20]                                        (tildex) at (3,  -1) {$\,\tilde{x}\mid z,w\,$};

% -------- directed edges (axis-aligned) -----------------------
\draw[->] (y)    -- (wxy);

\draw[->] (x)    -| (wxy);
\draw[->] (x)    -- (zx);

\draw[->] (zx)   -- (hatx);

\draw[line width=0.5pt] (hatx.east) ++(0.1,0) -- ++(0,-0.15) -- ++(0,0.3);
\draw[line width=0.5pt] (hatx.east) ++(0.1,0) -- (haty);

\draw[->] (zx)   -- (tildex);
\draw[->] (wxy)  |- (tildex);

\end{tikzpicture}
\caption{Encoder–decoder architecture: the inhibition arrow from $\hat{y}$ to $\hat{x}$ corresponds to the adversarial component.}
\label{fig:enc-dec}
\end{figure}

In order to optimize the objective in Equation \ref{eq:maxmin} with respect to $q_{z\vert x}, \, q_{w\vert x,y}$, and $g_{y\vert z}$ over the dataset $\mathcal{D}$, we introduce 
an encoder-decoder framework (illustrated in Figure \ref{fig:enc-dec}). In this framework, two separate reconstructions of $x$ are generated: one, denoted $\hat{x} \sim p_{x\vert z}$, where $z$ is sampled from the condition-invariant posterior $q_{z\vert x}$, and the other, denoted $\tilde{x} \sim p_{x \vert z, w}$, where in addition, $w$ is sampled from the condition-aware posterior $q_{w \vert x,y}$. The corresponding procedure is summarized in Algorithm \ref{alg}.

\textbf{Condition-agnostic representation \hspace{0.2em}} An input $x\in \mathcal{X}$ is mapped to the variational parameters $\theta_z = (\mu_{z}, \sigma^2_{z})$ by an encoder neural network
$f^z_\phi: \mathcal{X} \rightarrow \mathbb{R}^d \times \mathbb{R}^d_{+}$ parametrized by  weights $\phi$. A latent encoding $z \sim q_{\theta_z}$ is then sampled and mapped to a reconstruction $\hat{x}$ via a decoder neural network $h^z_\psi: \mathbb{R}^d \rightarrow \mathcal{X}$ parametrized by weights $\psi$.

\textbf{Adversarial classifier \hspace{0.2em}} Instead of training a high-capacity classifier directly from $z$ to $y$, we use the reconstruction $\hat{x}$ from $z$, and predict $y$ from $\hat{x}$ via a simpler model $g_{\beta}: \hat{\mathcal{X}} \rightarrow [0,1]^K$ (in most cases, implemented as a multinomial logistic regression with class-specific weights $\beta = {\beta_k}_{k=1}^K$, or a shallow MLP). 
Since $\hat{x}$ is a deterministic function of $z$, this is equivalent to applying a restricted classifier on $z$. By the data processing inequality, such a classifier can only capture a subset of the information $z$ contains about $y$; as a result, maximizing this lower bound on $I(z;y)$ also maximizes $I(z;y)$ itself.
Although this substitution weakens the estimation of the cross-entropy term $-\E_{q_{z \mid x}} \log g(y \mid z)$ from an information-theoretic standpoint, we observed this to be often advantageous in practice: predicting $y$ from $\hat{x}$ reduces the variance introduced by sampling $z \sim q_{\theta_z}$, providing a regularizing effect that prevents $q_{\theta_z}$ from overfitting to noisy classifier signals. We empirically evaluate this design choice in Appendix \ref{sec:ablations}, where we compare it to a classifier operating directly on $z$.

\textbf{Condition aware representation \hspace{0.2em}}  A labeled input pair $(x,y) \in \mathcal{X} \times \{1, \dots, K\}$ is mapped to the parameters $\theta_w = (\mu_{w}, \sigma^2_{w})$ using an encoder neural-network $f^w_\rho: \mathcal{X} \times \{1, \dots, K\} \rightarrow \mathbb{R}^d \times \mathbb{R}^d_{+}$ parametrized by weights $\rho$. A sample $w \sim q_{\theta_w}$ is then drawn, and the pair $(z,w)$ is mapped to a reconstruction $\tilde{x}$  via a decoder neural-network $h^{z,w}_\eta: \mathbb{R}^{d_z+d_w} \rightarrow \mathcal{X}$, parametrized by weights $\eta$. 
To compute the prior $p_{w\vert y}$, we estimate the class-specific mean as $\hat{\mu}_{k}\coloneqq\frac{1}{n_{k}}\sum_{i;y_i=k} z_{i}$ where each $z_i \sim q(z\mid x_i)$  is sampled from the encoder given an input $x_i$ with label $y_i = k$, and $n_k$ is the number of training points with the label $y=k$.

\begin{algorithm}[tb]
  \caption{}
  \label{alg}
  \begin{algorithmic}
    \STATE {\bfseries Input:} Data $\mathcal{D}=\{x_{1:n}, y_{1:n}\}$, number of training iterations $J$, initial parameters $\phi^{(0)}, \psi^{(0)}, \rho^{(0)}, \eta^{(0)}, \beta^{(0)}$, learning rates $\gamma_1, \gamma_2$, weighting terms $\alpha=(\alpha_1, \alpha_2)$
    \FOR{$1 \leq j \leq J$}
      \STATE Compute $\theta_z = f^z_{\phi^z}(x)$ and $\theta_w = f^w_{\rho^{(j-1)}}(x, y)$
      \STATE Sample condition invariant and aware latent variables $z \sim q_{\theta_z}$ and $w \sim q_{\theta_w}$
      \STATE Compute reconstructions $\hat{x} = h^z_{\psi^{(j-1)}}(z)$ and $\tilde{x} = h^{z,w}_{\eta^{(j-1)}}(z,w)$
      \STATE Compute condition prediction $\hat{y} = g_{\beta^{(j-1)}}(\hat{x})$
      \STATE Update classifier parameters:
      \begin{equation*}
        \beta^{(j)}
        \leftarrow 
        \beta^{(j-1)}
        - \gamma_1 \nabla_{\beta}\, \mathcal{L}_\alpha(q_{z \vert x}, q_{w \vert x,y}, g_{y \vert z})
        \end{equation*}
        with the gradient evaluated at 
        \begin{equation*}
            \Omega^{(j-1)} \coloneqq  \left(\phi^{(j-1)}, \psi^{(j-1)}, \rho^{(j-1)}, \eta^{(j-1)}\right).
        \end{equation*}
      \STATE Update encoder and decoder parameters $\Omega^{(j)}$
      \begin{equation*}
        \Omega^{(j)} 
        \leftarrow 
        \Omega^{(j-1)}
        + \gamma_2 \nabla_{\phi, \psi, \rho, \eta}\, \mathcal{L}_\alpha(q_{z \vert x}, q_{w \vert x,y}, g_{y \vert z})
      \end{equation*}
      with the gradient evaluated at $\beta^{(j)}$.
    \ENDFOR
    \STATE {\bfseries Return:} $\beta^{(J)}$, $\Omega^{(J)} = \left(\phi^{(J)},\psi^{(J)},\rho^{(J)},\eta^{(J)}\right)$.
  \end{algorithmic}
\end{algorithm}

In practice, the idealized game in Proposition~\ref{prop:optimal} is implemented with the standard relaxations used in VAE-based models. Specifically, we use a single-sample Monte Carlo estimate to approximate the expectations in Equation~\ref{eq:maxmin}. Instead of directly sampling from $q_\theta$, we employ the reparameterization trick to enable differentiable sampling: we sample $\epsilon \sim \mathcal{N}(0,I)$ and obtain a sample from $q_\theta$ by applying a deterministic transformation of $\epsilon$ based on the variational parameters $\theta$.

\section{Comparison to previous approaches} \label{sec:related}

Here we review VAE-based methods for disentangled representation learning, which form the primary basis for comparison with our approach. Broader related literature is discussed in Appendix~\ref{sup:related}.

VAEs \citep{vae} are generative models that learn latent representations by maximizing the evidence lower bound (ELBO) on the data log-likelihood: 
\begin{equation*}  \textstyle
\E_{q_{z\vert x}}[\log p(x\mid z)]-\KL\left(q_{z\mid x}\ \Vert \ p_{z}\right)\leq\log p(x),
\end{equation*}
where $(x,z)\sim p$, and $z\vert x \sim q$ is a latent variable inferred from data.  VAEs consist of an encoder $q_{z \mid x}$ that maps inputs to latent distributions, and a decoder $p_{x \mid z}$ that reconstructs inputs from latent representations. The learning process frames posterior inference as KL-regularized optimization over a variational family $\mathcal{Q}$, aiming to approximate the posterior $p_{z\vert x}$ under a typically simple prior $p(z)$.
Several VAE extensions were proposed to encourage disentanglement. These are discussed below.

\textbf{Conditional VAEs \hspace{0.2em}} \citep{CVAE} incorporate supervision into the standard VAE model by conditioning both the encoder and decoder on an observed label $y$, yielding the following objective:
\begin{equation*} \textstyle
    \E_{q_{z\vert x, y}} [\log p(x\vert z, y)]-\KL \left(q(z\vert  x, y) \Vert p(z)\right). 
    % \leq \log p(x \vert y).
\end{equation*}
While 
this allows controlled generation and partial disentanglement between $z$ and $y$,  since the prior $p(z)$ is global (e.g. $\mathcal{N}(0,I)$) and $z$ is inferred from both $x$ and $y$, no mechanism forces $z$ to discard label information. On the contrary, encoding both $x$-specific and $y$-specific information in $z$ will improve reconstruction error.

\textbf{Conditional Subspace VAEs \hspace{0.2em}}(CSVAEs) \citep{CSVAE},
explicitly factorize the latent space into a shared component $z$ and a label-specific component $w$ (see Supplementary Figure \ref{supfig:csvae_enc}). Similarly, their  hierarchical extension \citep{beker2024patches} introduces an intermediate latent variable between  $x$ and $(z, w)$.
As in our approach, to encourage disentanglement, CSVAEs introduce an adversarial regularization term that penalizes mutual information between $z$ and $y$, thereby discouraging predictability of $y$ from $z$. They are learned by optimizing
\begin{align*} 
     & \E_{q_{z,w\vert x,y}}[\log p(x\mid w,y)]\!-\!\E_{q_{z\vert x}}\!\left[\int \!g(y\mid z)\log g(y\mid z)\,dy\right]\!\\
    & -\!\KL\left(q_{w\mid x,y}\,\Vert\,q_{w\mid y}\right)\!-\!\KL\left(q_{z\mid x}\,\Vert\,p_{z}\right).
\end{align*}
However, here the reconstruction $p(x \mid w,y)$ uses only the condition-specific representation $w$, and therefore may result in uninformative invariant representation $z$. 

\textbf{Domain Invariant VAEs (DIVA) 
\hspace{0.2em}}  \citep{diva}, shown in Supplementary Figure \ref{supfig:diva_ccvae_enc},  
introduce two latent variables, $z$ and $w$, where $w$  captures label-related features by jointly optimizing a classifier $q(y \mid w)$ alongside the remaining objective.
For fully supervised cases, the DIVA model optimizes
\begin{align*} 
      & \E_{q_{z,w\vert x}}[\log p(x\mid z,w)]+\E_{q_{w\vert x}}[\log q(y\mid w)]\\
 & -\KL\left(q_{z\vert x}\Vert p_{z}\right)-\KL\left(q_{w\mid x}\Vert p_{w\mid y}\right).
 % \leq\log p(x,y).
\end{align*}
This objective corresponds to the assumptions $x \perp y \mid z,w$ and $z \perp w$ unconditionally, and therefore does not match the true posterior dependencies once conditioned on $x$. Furthermore, since the objective does not include an adversarial term acting on $z$, it may still encode information regarding $y$.

\textbf{ Characteristic-capturing VAEs (CCVAE) \hspace{0.2em}}  \citep{CCVAE} assume the same probabilistic model as DIVA (shown in Supplementary Figure \ref{supfig:diva_ccvae_enc}), but optimize a different objective,
\begin{align*} 
     & \E_{q_{z,w\vert x,y}}\left[\frac{q(y\mid w)}{q(y\mid x)}\log\frac{p(x\mid z,w)}{q(y\mid w)}\right]\\
    & -\KL\left(q_{z\vert x}\Vert p_{z\vert y}\right)-\KL\left(q_{w\mid x}\Vert p_{w\mid y}\right)+\log q(y\mid x).
\end{align*}
Here, the prior $p_{z \mid y}$ encodes the assumption that $z$ should depend on $y$, and therefore allows the invariant representation to in fact be condition-specific.

\textbf{Summary and comparison: \hspace{0.2em}} 
% In prior methods, reconstruction is performed jointly from both representations $z$ and $w$, via $p(x \mid z, w)$. This design provides no incentive to distribute information meaningfully between $z$ and $w$: the model can place all relevant information into $w$, leaving $z$ either uninformative or entangled with $w$.
Prior methods either lack an explicit mechanism forcing the shared latent $z$ to discard label information \citep{CVAE,diva}, reconstruct $x$ solely from the condition-specific representation, thereby allowing the invariant representation $z$ to remain uninformative \citep{CSVAE}, impose an unconditional independence assumption $z \perp w$ that does not match the true conditional independencies \citep{diva}, or encode in the prior that the invariant representation $z$ should in fact depend on $y$ \citep{CCVAE}.

Our method addresses these limitations by (i) optimizing a \emph{principled probabilistic objective} that enforces the correct conditional independencies, (ii) placing a \emph{prior over $w$ conditioned on the mean of $z$}, which keeps $z$ informative while discouraging leakage of information through label-specific aggregation, (iii) using \emph{two distinct reconstruction paths}, from the invariant and condition-specific representations, which further compel $z$ to capture informative shared structure, and (iv) incorporating an \emph{adversarial term} that penalizes leakage of class-specific information into the invariant representation $z$.

\section{Experiments}
\textbf{Datasets:}
We evaluate \methodname{} against existing approaches on synthetic data, natural images, and biological data. These datasets were chosen to probe condition-invariant structure and to ensure comparability with prior work. For instance, Swiss rolls and CelebA were used in \citet{CSVAE}, and CelebA also in \citet{CCVAE}. 

\textbf{Evaluation:}
When applicable, we evaluate reconstruction quality using negative log-likelihood (NLL), root mean squared error (RMSE), and the absolute deviation from the optimal-Bayes classifier on the reconstructed data, denoted as $\Delta$-Bayes.

Disentanglement is quantified via a neural estimator of the mutual information $I(z; w)$ \citep{pmlr-v80-belghazi18a}, and additional disentanglement metrics are reported in Appendix \ref{sec:metrics}.
Full model architectures, hyperparameters, and additional implementation details are provided in Appendix~\ref{sup:implement}.

Our results show that \methodname{} achieves superior performance across all experiments.

\subsection{Simulated data}
We begin with controlled synthetic experiments to isolate and visualize disentanglement.

\subsubsection{Parametric model} \label{res:1d}
\textbf{Data generating model:} Consider a model where the observed data $x$ is generated as a function of two latent variables $z$ and $w$, and $y$ is a binary label. Assume that the marginal distributions of the latent variables are given by $z\sim\mathcal{N}(0,1)$ and $w\sim\mathcal{N}(0,1)$, and that the data $x$ is generated as the sum of the two latent variables: $ x = z + w$. Since $z$ and $w$ are both drawn from $\mathcal{N}(0,1)$, it follows that $x \sim \mathcal{N}(0,2)$. Finally, assume that the binary label is determined by the sign of $w$: $y=1$ if $w > 0$, and $y=0$ otherwise.

\textbf{Optimal disentanglement:} Given that $z$ and $w$ are independent and $x=z+w$, we have that $p(z\mid x) = \mathcal{N}\!\left(z;\frac{x}{2},\frac{1}{2}\right)$. Hence,  given \(x\), the best estimate for \(z\) is $\frac{x}{2}$. Note that when ignoring the label $y$, the conditional  distribution $p(w\mid x)$ is $p(w\mid x) = \mathcal{N}\!\left(w;\frac{x}{2},\frac{1}{2}\right)$. However, the observation of $y$ (which indicates whether \(w\) is positive or negative) truncates this distribution: 
\begin{align*} \textstyle
     & p(w\vert x,y=1)\!=\!\frac{\mathcal{N}\!\left(w;\frac{x}{2},\frac{1}{2}\right)}{\Phi\!\left(\frac{x}{\sqrt{2}}\right)}, \; p(w\vert x,y=0)\!=\!\frac{\mathcal{N}\!\left(w;\frac{x}{2},\frac{1}{2}\right)}{1\!-\!\Phi\!\left(\frac{x}{\sqrt{2}}\right)}
\end{align*}

\textbf{Results:} 
Table \ref{table:1d_example} shows that \methodname\ (ours) best matches the analytic posteriors, yielding the lowest Bayes-classifier deviation and reconstruction error.

\begin{table*}[t]
    \centering
    \caption{Parametric model results: \methodname \ (ours) outperforms all competitors across all metrics.}
    \label{table:1d_example}
    % \small
    \begin{tabular}{lccccc}
    \\ \toprule
    & NLL $\downarrow$ & $\KL(q_{z \mid x} \;||\: p_{z\mid x}) \downarrow$ & $\KL(q_{w \mid x, y} \;||\: p_{w \mid x, y}) \downarrow$ & $\Delta$ -- Bayes $\downarrow$ \\
    \midrule
    CSVAE No Adv. & $1.810 \pm  0.016$ & $ 6.65 \pm  3.46$ & $ 23.61 \pm  0.36$ & $24.83 \pm  0.04$  \\
    CSVAE & $1.786 \pm  0.022$  & $ 2.85 \pm  1.11$ & $ 23.98 \pm  4.36$ & $24.33 \pm  1.28$ \\
    HCSVAE No Adv. & $1.784 \pm  0.010$ & $ 4.01 \pm  0.07$ & $ 25.82 \pm  0.38$ & $24.99 \pm  0.01$ \\
    HCSVAE & $1.770 \pm  0.004$ & $ 3.99 \pm  0.09$ & $ 26.25 \pm  0.59$ & $24.99 \pm  0.01$ \\
    DIVA & $1.788 \pm  0.008$ & $ 3.21 \pm  1.52$ & $ 12.88 \pm  3.31$ & $3.51 \pm  0.32$ \\
    CCVAE & $1.785 \pm  0.006$ & $ 1.77 \pm  0.81$ & $ 12.95 \pm  3.35$ & $3.57 \pm  0.15$\\
    
    \midrule
    \methodname\ (ours) & $\mathbf{1.769 \pm  0.003}$ & $ \mathbf{0.17 \pm  0.01}$ & $ \mathbf{10.10 \pm  0.73}$ & $ \mathbf{0.1 \pm  0.28}$ \\
    \bottomrule
    \end{tabular}
\end{table*}

\subsubsection{Noisy Swiss roll} \label{res:nsr}
\begin{figure}[!h]
    \centering
    \includegraphics[width=0.9\columnwidth]{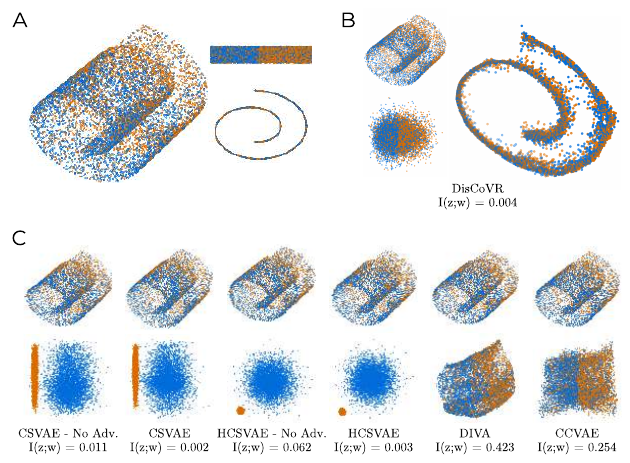}
    \caption{A: Noisy labeled Swiss Roll dataset. B: \methodname{} recovers the conditional embedding and reconstruction (left), while the shared embedding recovers the marginal spiral structure (right). C: Across models, ours best matches the disentangled ideal: $z$ captures the clean Swiss-roll geometry independently of the label, while label variation is isolated in $w$.}
    \label{fig:noisy_swiss_roll}
\end{figure}

\textbf{Dataset: \hspace{0.2em}}
We use a noisy variant of the labeled Swiss Roll dataset \citep{swissroll, CSVAE}, generating $n=20,000$ samples and assigning binary labels based on a lengthwise split, with labels flipped at rate $\rho$. The common geometry (its projection along the 2D plane) remains intact, while the conditional structure along the third axis becomes noisy. Figure \ref{fig:noisy_swiss_roll}A illustrates the setup.
\begin{table}[h]
  \centering
  \caption{Noisy Swiss roll results: \methodname{} \ (ours) yields lowest deviation from optimal-Bayes,  maintains low latent leakage, and high reconstruction accuracy.}
  \label{table:noisy_swiss_roll}  
  \setlength{\tabcolsep}{3pt}
  \small
  \resizebox{\linewidth}{!}{
  \begin{tabular}{lccclccc}
    \\ \toprule
    & $I(z;w) \downarrow$ & NLL $\downarrow$ & $\Delta$ -- Bayes $\downarrow$  \\
    \midrule
    CSVAE No Adv. & $0.047 \pm  0.023$ & $3.303 \pm  0.003$ & $23.88 \pm  12.02$ \\
    CSVAE & $0.031 \pm  0.025$  & $\mathbf{3.302 \pm  0.003}$ & $17.99 \pm  14.58$ \\
    HCSVAE No Adv. & $ 0.024 \pm  0.012$ & $\mathbf{3.302 \pm  0.002}$ & $30.00 \pm  0.00$ \\ 
    HCSVAE & $ \mathbf{0.002 \pm  0.001}$ & $\mathbf{3.302 \pm  0.002}$ & $30.00 \pm  0.00$ \\
    DIVA & $ 0.336 \pm  0.083$  & $\mathbf{3.302 \pm  0.003}$ & $1.88 \pm  1.05$\\
    CCVAE & $0.502 \pm  0.089$ & $\mathbf{3.302 \pm  0.002}$ & $2.21 \pm  0.84$ \\
    \midrule
    \methodname\ (ours) & $0.005 \pm  0.002$  & $\mathbf{3.302 \pm  0.002}$ & $\mathbf{1.14 \pm  0.21}$\\ 
    \bottomrule
  \end{tabular}
   }
\end{table}

\textbf{Optimal disentanglement: \hspace{0.2em}} 
Since the Swiss Roll is sliced at the center and label noise is applied uniformly, marginalizing over labels yields a symmetric spiral centered along the roll—i.e., the marginal posterior $p(z \mid x)$ is label-invariant. In contrast, the conditional component retains a noisy but informative signal, with a uniform noise level of $\rho=0.3$.  As a result, the Bayes optimal classifier trained on any realistic representation is upper-bounded at 70\% accuracy. 

\textbf{Results: \hspace{0.2em}}
Figure~\ref{fig:noisy_swiss_roll} presents qualitative and quantitative results, showing that \methodname{} both models the label noise accurately and effectively disentangles shared and condition-specific structure. Notably, \methodname{} captures the marginal data distribution, successfully recovering the expected spiral pattern, as shown in Figure~\ref{fig:noisy_swiss_roll}B.

Additionally, the results in Table \ref{table:noisy_swiss_roll} show that \methodname\ achieves the lowest deviation from the optimal Bayes classifier and minimal information leakage between latent variables, while preserving reconstruction quality. This confirms that label information is concentrated in $w$ while $z$ remains both informative and label-invariant.

\subsection{Real data}

\subsubsection{Noisy colored MNIST} \label{res:ncmnist}
\textbf{Dataset: \hspace{0.2em}} We construct a modified MNIST \citep{deng2012mnist} dataset from $60,000$ duplicated images: in one copy we remove the red channel ($y=0$) and in the other we remove the green channel ($y=1$), so that the digit shape remains intact and is carried entirely by the blue channel. Label noise is introduced by flipping $y$ with probability $\rho \in \{0, 0.1, 0.2, 0.3, 0.4\}$. 

\begin{figure}[h]
    \centering
    \includegraphics[width=0.9\columnwidth]{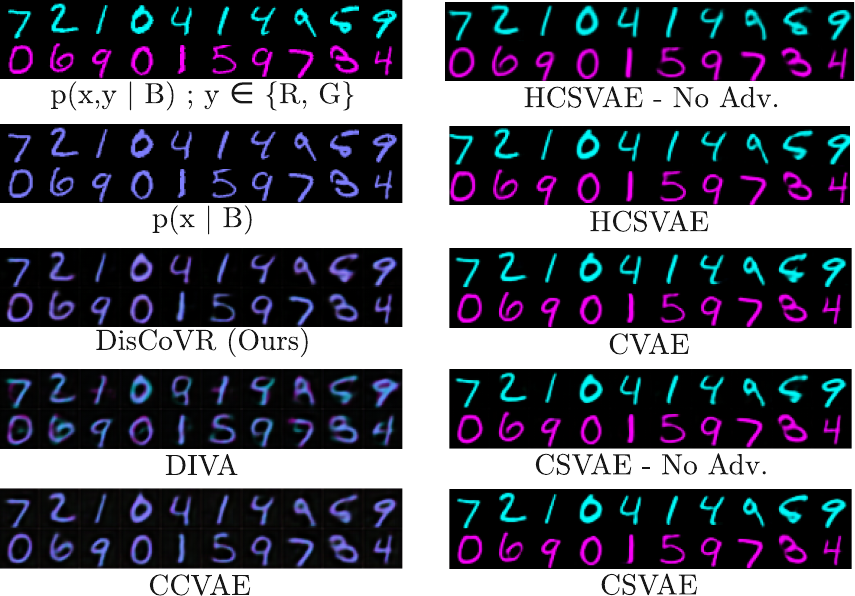}
    \caption{Colored MNIST reconstructions from the label-agnostic representation $z$ at noise level $\rho=0.3$. Only \methodname{} consistently produces mixed semi-red/blue (purple) digits, indicating that color information has been removed from $z$ and that the reconstructions approximate the true marginal.}
    \label{fig:cmnist}
\end{figure}

\textbf{Optimal disentanglement: \hspace{0.2em}} With colors balanced across labels, the ideal $z$-reconstruction averages colors over labels, retaining one mixed color (Figure \ref{fig:cmnist}).

\textbf{Results: \hspace{0.2em}} We evaluate marginal coloring reconstruction by \methodname{} and previous methods. Under no label noise ($\rho = 0$), all methods perform similarly (see Supplementary Figure~\ref{supfig:cmnist_nonoise}). 

However, at all non-zero noise levels, \methodname{} consistently outperforms competing methods and is the only approach that reconstructs digits in purple, correctly averaging over the two colors.

Metrics for $\rho = 0.3$ are shown in Supplementary Table~\ref{table:noisy_cmnist}, with results for other noise levels in Supplementary Figure~\ref{supfig:cmnist_noises}.       

\subsubsection{CelebA} \label{res:celeba}
\textbf{Glasses attribute: \hspace{0.2em}} We use all CelebA  \citep{liu2015faceattributes} images labeled with  \emph{eyeglasses} attribute ($y=1$), and twice as many randomly sampled images without glasses ($y=0$), resulting in $n=35,712$ images in total.

\textbf{Hat attribute: \hspace{0.2em}} Results for an analogous experiment with the wearing-hat attribute are provided in Appendix \ref{sup:hats}.

\begin{figure}[H]
    \centering
    \includegraphics[width=\columnwidth]{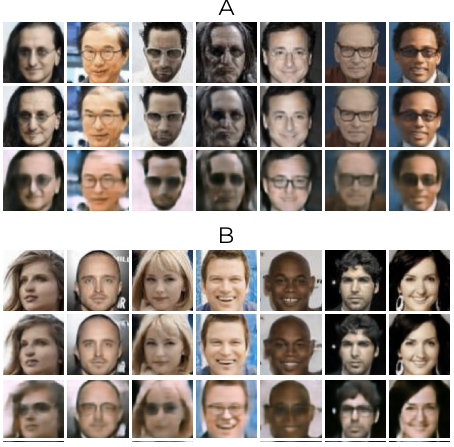}
    \caption{CelebA-Eyeglasses results. Top: Original images with (A) or without eyeglasses (B). Middle: Full reconstructions by \methodname. Bottom: reconstructions solely from invariant embeddings $z$. The condition-specific representation needs to be invariant to $y$ (presence or absence of glasses). Indeed, all reconstructed faces display an intermediate "pseudo-glasses" appearance in both A and B, regardless of their presence in the original images.}
    \label{fig:celeba}
\end{figure}

\textbf{Results: \hspace{0.2em}} Figure~\ref{fig:celeba} shows that \methodname{} accurately reconstructs input images while producing shared embeddings that marginalize over the \textit{eyeglasses} attribute, consistently adding "pseudo-glasses" to all samples. 

Competing methods are shown in Supplementary Figure~\ref{supfig:celeba_recons}, with quantitative results in Supplementary Table~\ref{suptable:celeba-res}. 

While full reconstruction quality from both representations together is comparable across methods, \methodname{} achieves notably better disentanglement. In this experiment, however, the adversarial classifier incurs higher computational cost compared to other methods. See Table \ref{suptable:time-epoch} for additional details.

\subsubsection{Single cell RNA-sequencing} \label{res:kang}

\begin{figure}[ht]
    \centering
    \includegraphics[width=\columnwidth]{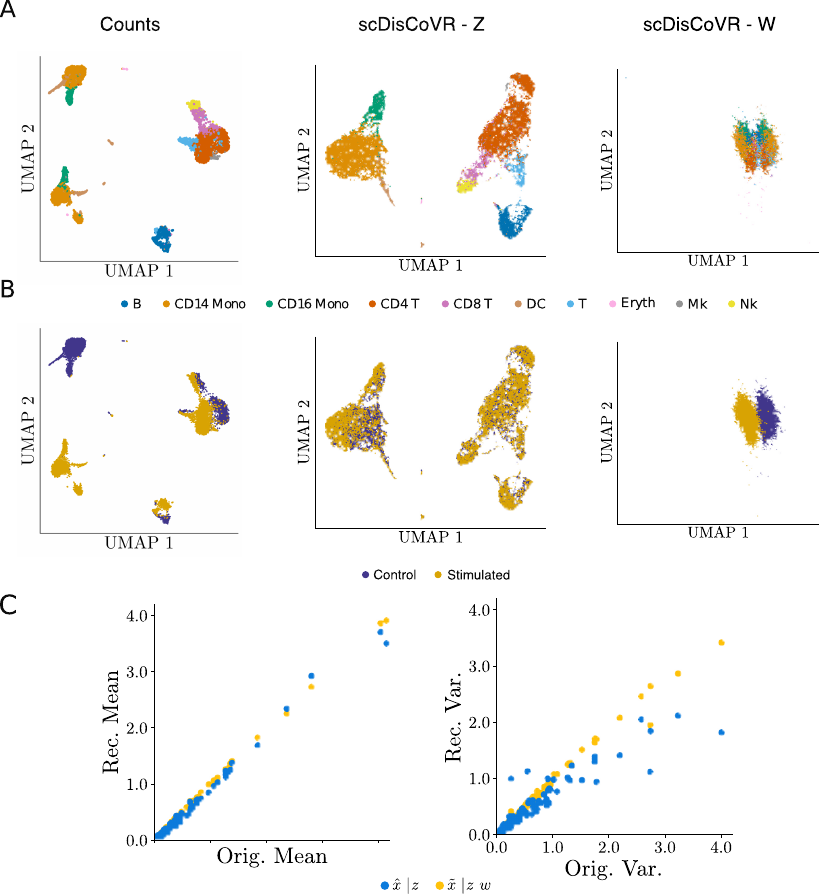}
    \caption{A–B (left): UMAPs of raw gene counts from the IFN-$\beta$ dataset. A–B (middle): Shared embedding $z$ aligns cells by type while removing stimulation effects. A–B (right): Condition-specific embedding $w$ isolates the stimulation effect. C: Reconstructions from both $z$ and $w$ (yellow) recover empirical gene means and variances, while reconstructions from $z$ alone (blue) miss the stimulation-induced variance, confirming that $z$ discards $y$ while preserving cell-type features.
   }  
    \label{fig:kang}
\end{figure}

\textbf{Dataset: \hspace{0.2em}} 
We analyze single-cell RNA sequencing from $n = 13,999$ peripheral blood mononuclear cells (PBMCs) collected from 8 lupus patients under two conditions:  7,451 cells control ($y=0$), and 6,548 IFN-$\beta$ stimulation cells. IFN-$\beta$ stimulation induces notable shifts in gene expression, visible in the UMAP embedding in Figure \ref{fig:kang}B (left).

\textbf{Results: \hspace{0.2em}}
 Supplementary Table \ref{suptable:kangclust} shows that \methodname\ effectively achieves the desired behavior with strong empirical performance, where only cell type information is captured in $z$ (Figure \ref{fig:kang}A, middle) while the effects of IFN-$\beta$ stimulation are wholly represented in $w$ (Figure \ref{fig:kang}B, right). Other approaches either (1) achieve mixing in the $z$ space, but compromise on keeping cell types separated or (2) leak information about stimulation into the $z$ space (Supplementary Figure \ref{supfig:kang_others}).   

\textbf{Facilitating interpretability: \hspace{0.2em}} By enabling marginalized reconstructions, \methodname{} provides a direct link between shared embeddings and gene expression, offering clearer insight into the effects of IFN-$\beta$ stimulation, unlike other methods. In Figure \ref{fig:kang}C, comparing variance across marginal and full reconstructions accurately recovers gene-level differences associated with IFN-$\beta$ stimulation, including \emph{ISG15}, \emph{FTL}, \emph{CCL8}, \emph{CXCL10}, \emph{CXCL11}, \emph{APOBEC3A}, \emph{IL1RN}, \emph{IFITM3} and \emph{RSAD2}.

\section{Conclusion} \label{conc}
In this work we introduced a variational framework for disentangled representation learning in multi-condition datasets that explicitly separates condition-invariant and condition-specific factors. Unlike prior work, \methodname{} is built around a principled probabilistic objective that encodes the correct conditional independencies, a prior over $w$ conditioned on the class-wise mean of $z$, and an adversarial term that limits label information in the invariant representation. The model uses two reconstruction paths, which forces $z$ to remain informative about shared structure.

Across synthetic benchmarks and real-world datasets, \methodname{} achieves strong reconstruction, low information leakage, and accurate modeling of conditional effects, consistently outperforming existing methods in disentangling shared and condition-specific structure.

\section*{Acknowledgements}
BD acknowledges the support of the CIFAR MacMillan Multiscale Human Project and the National Institute of General Medical Sciences of the National Institutes of Health under award number R35 GM157082-02.
DB acknowledges the support of NSF IIS-2127869, NSF DMS-2311108, ONR N000142412243, and the Simons Foundation.
YS acknowledges the support of the Founder’s Postdoctoral Fellowship, Department of Statistics, Columbia University.

\section*{Impact Statement}
This paper presents work whose goal is to advance the field of Machine
Learning. There are many potential societal consequences of our work, none
which we feel must be specifically highlighted here.

% \clearpage
\bibliography{references}
\bibliographystyle{icml2026}

%%%%%%%%%%%%%%%%%%%%%%%%%%%%%%%%%%%%%%%%%%%%%%%%%%%%%%%%%%%%%%%%%%%%%%%%%%%%%%%
%%%%%%%%%%%%%%%%%%%%%%%%%%%%%%%%%%%%%%%%%%%%%%%%%%%%%%%%%%%%%%%%%%%%%%%%%%%%%%%
% APPENDIX
%%%%%%%%%%%%%%%%%%%%%%%%%%%%%%%%%%%%%%%%%%%%%%%%%%%%%%%%%%%%%%%%%%%%%%%%%%%%%%%
%%%%%%%%%%%%%%%%%%%%%%%%%%%%%%%%%%%%%%%%%%%%%%%%%%%%%%%%%%%%%%%%%%%%%%%%%%%%%%%
\newpage
\appendix
\onecolumn
\section{Additional related work} \label{sup:related}

\subsection{Domain generalization}

The task of representation disentanglement is closely related to the field of domain generalization \citep{muandet2013domain}, which assumes limited or no access to target domain samples and aims to learn representations that can be readily adapted, often via transfer learning, to new, unseen domains.

As noted by \citet{wang2019learning}, existing methods in domain generalization can be broadly categorized into two main approaches: (i) approaches for reducing the inter-domain differences, often by using adversarial techniques \citep{ghifary2015domain, wang2017select, motiian2017unified, li2018domain, carlucci2019,  wang2019learning, akuzawa2020adversarial, zhu2022localized, gokhale2023improving, dayal2023madg, cheng2023adversarial, chen2024domain}, and (ii) Approaches that construct an ensemble of domain-specific models, and then fuse their representations to form a unified, domain-agnostic representation \citep{ding2017deep, mancini2018best, zhou2021domain, muhammad2024domain}.

Additional strategies for domain generalization include contrastive learning approaches \citep{kim2021selfreg}, methods based on distribution alignment via metrics
\citep{muandet2013domain, sun2016deep}, and techniques utilizing custom network architectures, for instance by incorporating domain-specific adapters between shared layers \citep{rebuffi2017learning, rebuffi2018efficient, li2019efficient, omi2022model}.

The primary distinction between these methods and ours lies in the explicit probabilistic modeling and disentanglement of domain-invariant and domain-specific factors. Whereas prior approaches typically focus on aligning domains through adversarial training or fusing multiple domain-specific predictors, our method constructs a structured latent space, decomposed into a condition-specific representation $z$, capturing domain-invariant information, and a conditional component 
$w$, which encodes domain-specific variability. This factorization is learned through a tailored variational objective involving an adversarial penalty and two reconstructions —one based on 
$z$ alone, and another on the full latent pair 
$(z,w)$, thereby promoting both interpretability and a clean separation of shared and domain-aware features.

\subsection{Out of distribution generalization}
\subsubsection{Environment balancing methods}
The field of out-of-distribution (OOD) generalization emerged from foundational work on causality and invariance across training environments \citep{peters2016causal, peters2017elements}. The central assumption is that each environment exhibits distinct spurious correlations between features and labels; therefore, robust generalization requires models to focus on invariant relationships that hold across environments. To address this distribution shift, many recent approaches adopt a regularized empirical risk minimization framework:
\begin{equation} \label{eq:ood_general}
    \min_\theta \quad \smashoperator{\sum_{e \in E_\text{train}}} \ell^e(f_\theta) + \lambda R(f_\theta, E_\text{train}),
\end{equation}
where the regularizer $R$ encourages representations that are stable across environments. Among these, Invariant Risk Minimization (IRM) \citep{arjovsky2019invariant} enforces that a single classifier remains optimal across all environments, Variance Risk Extrapolation (VarREx) \citep{krueger2021out} promotes robustness by minimizing the variance of losses across environments, and  CLOvE \citep{wald2021calibration} takes a calibration-theoretic perspective, penalizing discrepancies between predicted confidence and correctness across environments.

While these methods focus on enforcing predictive invariance across environments through regularization, our approach instead explicitly enforces conditional independence between the shared latent variable $z$ and an environment-aware variable $w$.

\subsubsection{Distributionally robust methods}

An alternative line of work for handling distribution shifts is Distributionally Robust Optimization (DRO) \citep{DRO, duchi2021statistics, duchi2021learning, wei2023distributionally}, which avoids assuming a fixed data-generating distribution. Instead, DRO methods optimize performance under the worst-case scenario over a family of plausible distributions. A prominent variant, known as group DRO \citep{sagawa2019distributionally, piratla2021focus}, introduces group-level structure that may correlate with spurious features, potentially leading to biased predictions. In settings where group labels are not directly observed, several strategies have been proposed, including reweighting high-loss examples \citep{liu2021just} and balancing class-group combinations through data sub-sampling \citep{idrissi2022simple}.

However, these approaches assume that the label space remains fixed between training and test time, limiting their applicability in adaptation to new domains, environments or conditions.

\subsection{Zero-shot learning}

Zero-shot learning systems \citep{few_shot, zero_shot} aim to classify instances from novel, previously unseen classes at test time. In contrast to the out-of-distribution (OOD) generalization setting, these approaches typically do not assume the presence or structure of a distribution shift. Instead, a common strategy is to learn data representations that capture class-agnostic similarity, enabling the model to determine whether two instances belong to the same class without requiring knowledge of the class identity itself. Such methods include contrastive-learning \citep{contrastive}, siamese neural networks \citep{siamese_networks},  triplet networks \citep{triplet_networks}, and other more recent variations \citep{oh2016deep, sohn2016improved, wu2017sampling, yuan2019signal}. 
Recent work has begun to address the impact of class distribution shifts in zero-shot settings. For instance, \citet{slavutsky2024class} integrate environment-based regularization—motivated by OOD generalization—with zero-shot learning by simulating distribution shifts through hierarchical sampling, enabling the model to learn representations that are robust to shifts in class distributions.

While this line of work shares our motivation of improving robustness under unseen conditions, it primarily addresses the problem of class-level generalization through similarity-based learning, rather than explicitly modeling and disentangling the latent factors—such as domain or environment—that drive distributional variation across tasks.

\section{Proofs} \label{sup:proofs}
\subsection{Proof of Proposition \ref{prop:obj_gap}} \label{sup:obj_gap}

\begin{proof}

\begin{align}
 & \ELBO(q,p;x,y)-\mathcal{L}_{w}(q_{w\vert x,y},p;x,y)\\
 & =\left[\log p(x\mid y)-\KL(q_{w\vert x,y}\,\Vert\,p_{w\vert x,y})\right]-\left[\log p(x\mid y)-\KL\left(q_{z\vert x}q_{w\vert x,y}\,\Vert\,p_{z,w\vert x,y}\right)\right]\\
 & =\KL\left(q_{z\vert x}q_{w\vert x,y}\,\Vert\,p_{z,w\vert x,y}\right)-\KL\left(q_{w\vert x,y}\,\Vert\,p_{w\vert x,y}\right)\\
 & =\E_{q_{w\vert x,y}}\left[\E_{q_{z\vert x}}\left[\log q(z\mid x)+\log q(w\vert x,y)-\log p(z,w\mid x,y)\right]\right]\\
 & \phantom{=}-\E_{q_{w\vert x,y}}\left[\log q(w\vert x,y)-\log p(w\vert x,y)\right]\\
 & =\E_{q_{w\vert x,y}}\left[\E_{q_{z\vert x}}\left[\log q(z\mid x)-\log p(z,w\mid x,y)+\log p(w\vert x,y)\right]\right]\\
 & =\E_{q_{w\vert x,y}}\left[\E_{q_{z\vert x}}\left[\log q(z\mid x)-\log p(z\mid w, x,y)\right]\right]\\
 & =\E_{q_{w\vert x,y}}\left[\text{KL}\left(q_{z\vert x}\,\Vert\,p_{z\vert w,x,y}\right)\right].
\end{align}
\end{proof}

\subsection{Game equilibrium} \label{sup:game}

\subsubsection{Regularity conditions} \label{sup:reg}

To ensure that expectations and KL‑terms in the game objective $\mathcal{L}(q_{z\vert x}, q_{w\vert x, y}, g_{y\vert z} )$ render the functionals strictly concave in $q_{z\vert x}$ , strictly concave in $q_{w\vert x, y}$, and strictly convex in $g$, the following regularity conditions are required:

\begin{enumerate}
    \item The likelihoods $p(x\vert z), p(x \vert z, w),p(y \vert x)$ are strictly positive, continuous densities.
    \item The variational families $Q_z$ and $Q_w$, and the set of achievable classifiers $\mathcal{G}$ are non‑empty, convex and compact.
    \item $\log p(x\vert z,w)$ and $\log g(y \vert x)$ are integrable.
\end{enumerate}

\subsubsection{Proof of Proposition \ref{prop:optimal}} \label{sup:opt}

\begin{proof}

Since $\mathcal{L}_z(q_{z\vert x},p;x)$ is the standard ELBO objective, we have that
\begin{equation}
    \mathcal{L}_z(q_{z\vert x},p;x)=\log p(x)-\KL\left(q_{z \vert x}\,\Vert\,p_{z \vert x}\right).
\end{equation}
Similarly, we have that
\begin{align}
\mathcal{L}_w(q_{w\vert x,y},p;x,y) & =\log p(x\mid y)-\KL\left(q_{z\vert x}q_{w\vert x,y}\,\Vert\,p_{z,w\vert x,y}\right).
\end{align}
Thus,
\begin{align}
\mathcal{L}(q_{z\vert x},q_{w\vert x,y},g_{y\vert z})=\E_{p_{x,y}} & \left[\log p(x)-\KL\left(q_{z\vert x}\,\Vert\,p_{z\vert x}\right)\right.\\
 & +\log p(x\mid y)-\KL\left(q_{z\vert x}q_{w\vert x,y}\,\Vert\,p_{z,w\vert x,y}\right)\\
 & \left. -\E_{q_{z\vert x}}\log g(y\mid z)\right].
\end{align}

For fixed $q_{z|x}$, the adversarial classifier minimizes:
\begin{equation}
    -\E_{p_{x,y}}\E_{q_{z\vert x}}\log g(y\mid z),
\end{equation}
which is the population cross-entropy and is strictly convex in $g(y\vert z)$, and thus has a unique solution.

It remains to show that the terms in the objective function that depend on $q_{z\vert x}$ and $q_{w\vert x,y}$, are strictly concave in each argument when the others are held fixed.

Focusing on the terms dependent on $q_{w\vert x,y}$ first, define
\begin{align}
\ell_{w}\coloneqq & -\KL\left(q_{z\vert x}q_{w\vert x,y}\,\Vert\,p_{z,w\vert x,y}\right)\\ \notag
= & -\iint q\left(z\mid x\right)q(w \mid x,y)\left[\log q\left(z\mid x\right)+\log q(w \mid x,y)-\log p(z,w\mid x,y)\right]\,dz\,dw\\
= & -\int q\left(z\mid x\right)\log q\left(z\mid x\right)\,dz-\int q(w \mid x,y)\log q(w \mid x,y)\,dw\\
 & +\iint q\left(z\mid x\right)q(w \mid x,y)\log p(z,w\mid x,y)\,dz\,dw\\
= &\label{eq:lw}  H(q_{z\mid x})+H(q_{w\mid x,y})+\E_{q_{z\vert x}}\E_{q_{w\vert x,y}}\log p(z,w\mid x,y).
\end{align} 
Note that 
\begin{equation}
    \E_{q_{z\vert x}}\E_{q_{w\vert x,y}}\log p(z,w\mid x,y)
\end{equation}
is linear in $q_{w\vert x,y}$, and since $H(q_{w\mid x,y})$
is strictly concave in $q_{w\vert x,y}$, we have that $\E_{p_{x,y}}[\ell_{w}]$
is strictly concave in $q_{w\vert x,y}$.

Similarly, define 
\begin{align}
\ell_{z} & \coloneqq-\KL\left(q_{z\vert x}\,\Vert\,p_{z\vert x}\right)-\KL\left(q_{z\vert x}q_{w\vert x,y}\,\Vert\,p_{z,w\vert x,y}\right).
\end{align}
By convexity of KL divergence in its first argument, $\ensuremath{-\KL\left(q_{z\vert x}\,\Vert\,p_{z\vert x}\right)}$
is strictly concave in $q_{z\mid x}$.

Focusing on the second KL term, from Equation \ref{eq:lw} we have that
\begin{equation}
    -\KL\left(q_{z\vert x}q_{w\vert x,y}\,\Vert\,p_{z,w\vert x,y}\right) = H(q_{z\mid x})+H(q_{w\mid x,y})+\E_{q_{z\vert x}}\E_{q_{w\vert x,y}}\log p(z,w\mid x,y),
\end{equation}
where $H(q_{z\mid x})$
is strictly concave in $q_{z\vert x}$. 

Recall that we assumed that $p(w\vert y)$ depends on $q(z\vert x)$.
Under our model 
\begin{equation}
    p(x,y,z,w)=p(y)p(w\mid y)p(z)p(x\mid z,w),
\end{equation}
yielding 
\begin{equation}
    p\left(z,w\mid x,y\right)=p(w\mid y)\frac{p(z)p(x\mid z,w)}{p(x\mid y)}.
\end{equation}
Hence,
\begin{align} \notag
\E_{q_{z\vert x}}\E_{q_{w\vert x,y}}\log p(z,w\mid x,y)=\E_{q_{z\vert x}}\E_{q_{w\vert x,y}}\left[\log p(w\mid y)+\log\frac{p(z)p(x\mid z,w)}{p(x\mid y)}\right],
\end{align}
where  $ p(w\mid y)=\mathcal{N}\left(w;\mu_{y},I\right)$
with $\mu_{y}=\E_{p_{x\vert y}}\left[\E_{q_{z\vert x}}[z]\right]$.
Therefore,
\begin{align}
 & \E_{q_{z\vert x}}\E_{q_{w\vert x,y}}\left[\log p(w\mid y)\right] =-\frac{1}{2} \left[d\log(2\pi)+\E_{q_{z\vert x}}\E_{q_{w\vert x,y}}\left\Vert w-\mu_{y}\right\Vert ^{2}\right]
\end{align}
where $-\left\Vert w-\mu_{y}\right\Vert ^{2}$ is a quadratic form
in $\mu_{y}$, which is linear in $q_{z\vert x}$, and thus $\E_{q_{z\vert x}}\E_{q_{w\vert x,y}}\left[\log p(w\mid y)\right]$
is strictly concave in $q_{z\vert x}$. Hence, $-\KL\left(q_{z\vert x}q_{w\vert x,y}\,\Vert\,p_{z,w\vert x,y}\right)$
is strictly concave in $q_{z\vert x}$, and thus so is $\E_{p_{x,y}}[\ell_z]$.
\end{proof}

%%%%%%%%%%%%%%%%%%%%%%%%%%%%%%%%%%%%%%%%%%%%%%%%%

\section{Supplementary Figures}

\renewcommand{\figurename}{Supplementary Figure}
\setcounter{figure}{0}

\begin{figure}[H]
\centering

% ---  Left sub‑figure ------------------------------------------
\begin{subfigure}[t]{0.48\textwidth}
\centering
\begin{tikzpicture}          
\tikzset{
  circ/.style={circle, draw=black, fill=gray!20, minimum size=6mm, inner sep=0pt},
  sqr/.style={rectangle, draw=black, minimum height=6mm, inner sep=1pt},
  wide/.style={minimum width=6mm}    
}

\node[circ] (y) at (0, 0) {$y$};
\node[sqr, wide] (wxy) at (1.5, 0) {$\,w\mid x,y\,$};
\node[sqr, wide] (zx) at (3, 0) {$\,z\mid x\,$};
\node[sqr, wide, draw=gray!60!black, line width=0.8pt] (haty) at (4.5, 0) {$\hat{y}$};
\node[circ] (x) at (3, 1) {$x$};
\node[sqr, wide, fill=gray!20] (tildex) at (3, -1) {$\,\tilde{x}\mid z,w\,$};

\draw[->] (y) -- (wxy);
\draw[->] (x) -| (wxy);
\draw[->] (x) -- (zx);
\draw[line width=0.5pt] (zx.east) ++(0.1,0) -- ++(0,-0.15) -- ++(0,0.3);
\draw[line width=0.5pt] (zx.east) ++(0.1,0) -- (haty);
\draw[->] (zx) -- (tildex);
\draw[->] (wxy) |- (tildex);

\end{tikzpicture}
\caption{}
\label{supfig:csvae_enc}
\end{subfigure}
\hfill
% --- Right sub‑figure ------------------------------------------
\begin{subfigure}[t]{0.48\textwidth}
\centering
\begin{tikzpicture}                    
\tikzset{
  circ/.style={circle, draw=black, fill=gray!20, minimum size=6mm, inner sep=0pt},
  sqr/.style={rectangle, draw=black, minimum height=6mm, inner sep=1pt},
  wide/.style={minimum width=6mm}    
}

\node[sqr, wide] (wx) at (1.5, 0) {$\,w\mid x\,$};
\node[sqr, wide] (zx) at (3, 0) {$\,z\mid x\,$};
\node[sqr, wide, draw=gray!60!black, line width=0.8pt] (haty) at (0, 0) {$\hat{y}$};
\node[circ] (x) at (3, 1) {$x$};
\node[sqr, wide, fill=gray!20] (tildex) at (3, -1) {$\,\tilde{x}\mid z,w\,$};

\draw[->] (wx) -- (haty);
\draw[->] (x) -| (wx);
\draw[->] (x) -- (zx);
\draw[->] (zx) -- (tildex);
\draw[->] (wx) |- (tildex);

\end{tikzpicture}
\caption{}
\label{supfig:diva_ccvae_enc}
\end{subfigure}

\caption{Encoder-decoder structures for previous approaches. (a) CSVAE. (b) DIVA - CCVAE.}
\label{supfig:enc_structures}
\end{figure}

\begin{figure}[H]
    \centering
    \includegraphics[width=\textwidth]{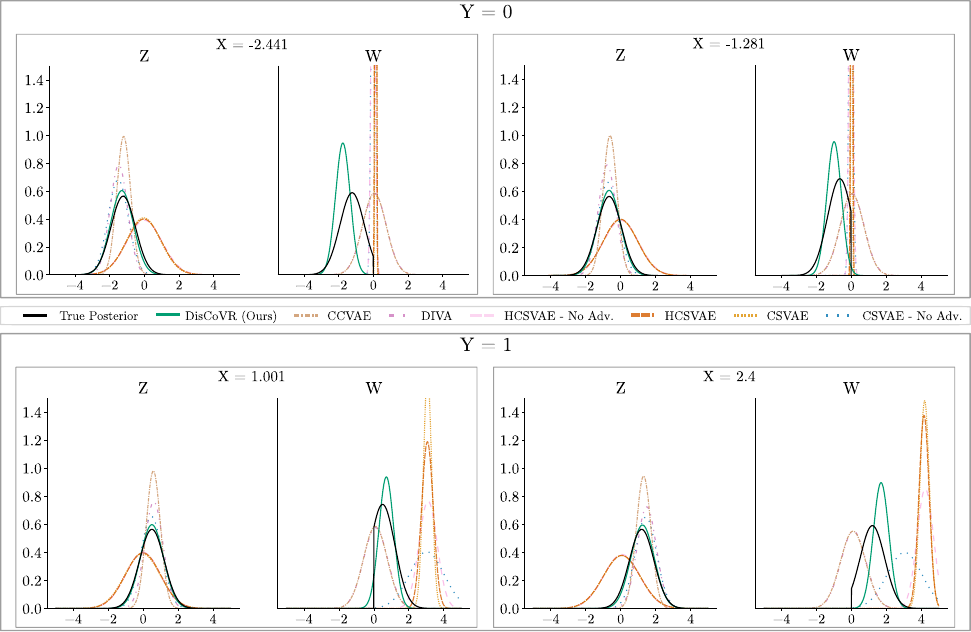}
    \caption{Comparison of approximate variational posteriors against the true posterior for latent variables $z, w$ for different values of $x$ with $y=0$ (top) and $y=1$ (bottom).}
    \label{supfig:1d}
\end{figure}

\begin{figure}[H]
    \centering
    \includegraphics[width=\textwidth]{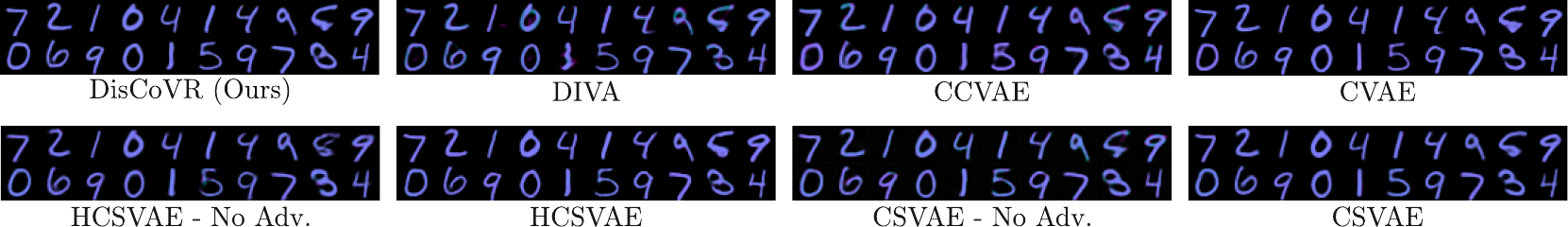}
    \caption{Colored MNIST results for no noise.}
    \label{supfig:cmnist_nonoise}
\end{figure}

\begin{figure}[H]
    \centering
    \includegraphics[width=\textwidth]{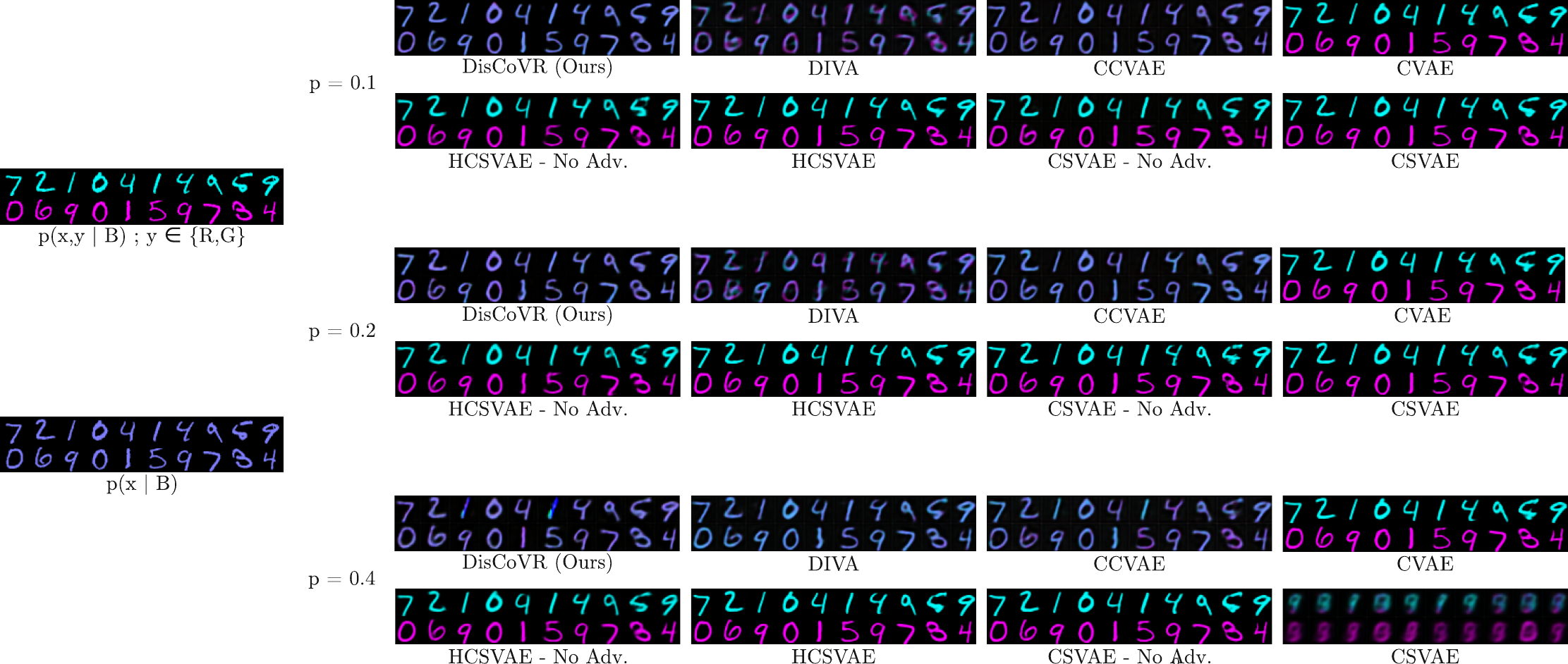}
    \caption{Colored MNIST visual results across the remaining noise levels.}
    \label{supfig:cmnist_noises}
\end{figure}

\begin{figure}[H]
    \centering
    \includegraphics[width=\textwidth]{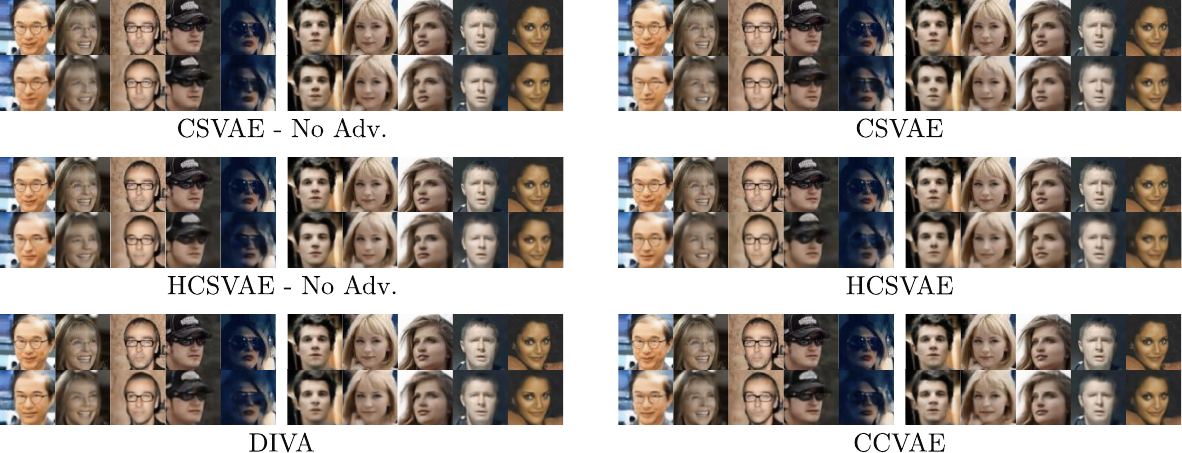}
    \caption{Reconstruction performance for other models on the CelebA-Glasses dataset. Top: Original samples from the data. Bottom: Reconstructions by the given model.}
    \label{supfig:celeba_recons}
\end{figure}

\begin{figure}[H]
    \centering
    \includegraphics[width=\textwidth]{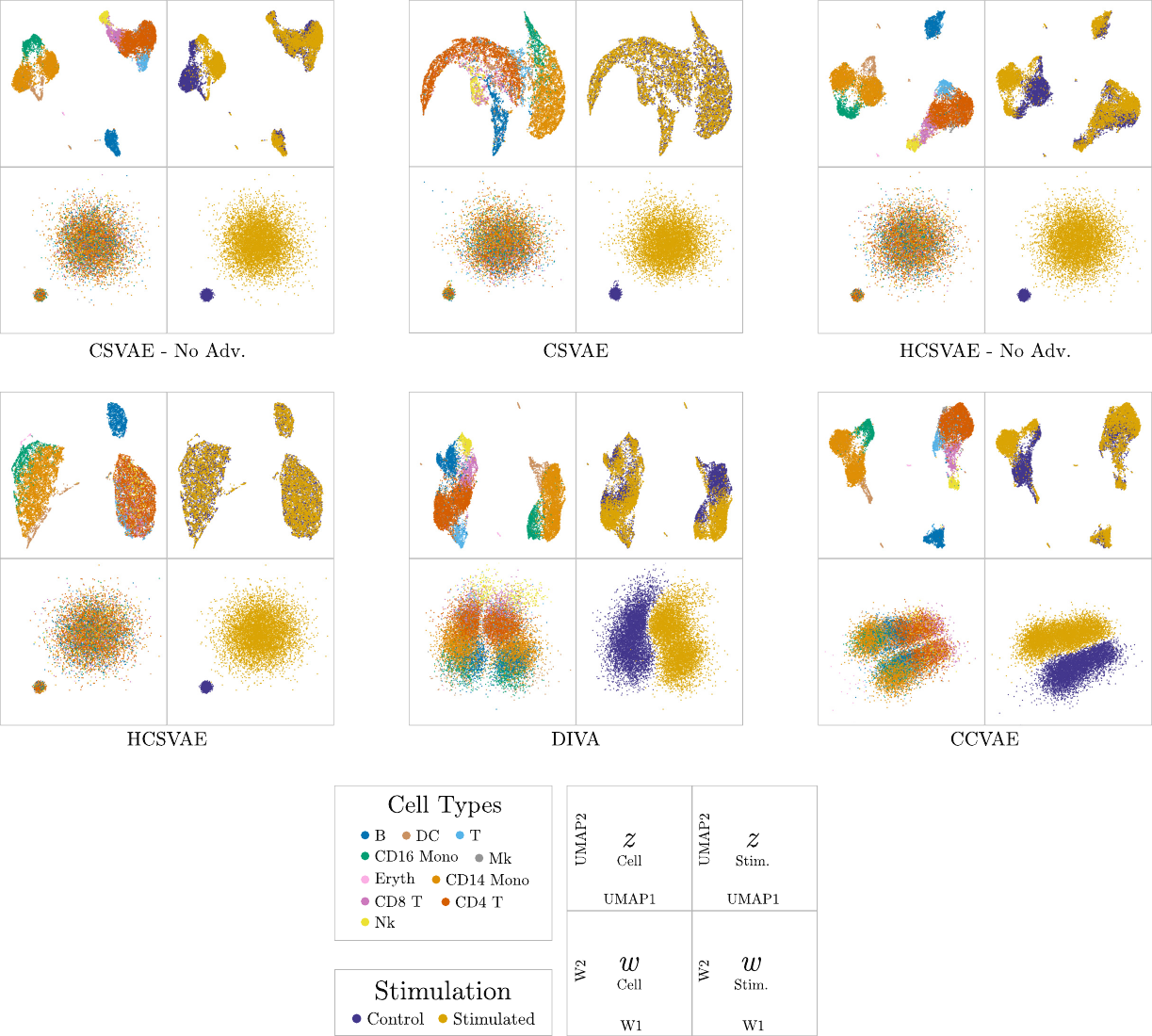}
    \caption{Embeddings obtained by other models on the Kang dataset. For each block, top (resp. bottom) rows are $z$ (resp. $w$) embeddings, while left (resp. right) columns are colored by cell type (resp. stimulation). }
    \label{supfig:kang_others}
\end{figure}

%%%%%%%%%%%%%%%%%%%%%%%%%%%%%%%%%%%%%%%%%%%%%
\section{Supplementary tables for experimental results}
\renewcommand{\tablename}{Supplementary Table}
\setcounter{table}{0}

\begin{table}[H]
  \caption{RMSE for the Colored MNIST dataset without any label noise.}
  \label{suptable:cmnist_no_noise}
  \centering
  \begin{tabular}{lc}
    \\ \toprule
    & {Marginal RMSE ($p=0$) $\downarrow$} \\
    \midrule
    CSVAE - No Adv. & $\mathbf{0.064 \pm 0.002}$ \\
    CSVAE & $0.079 \pm 0.008$  \\
    HCSVAE - No Adv. & $0.094 \pm 0.004$ \\
    HCSVAE & $0.079 \pm 0.030$ \\
    DIVA & $0.065 \pm 0.005$  \\
    CCVAE & $0.065 \pm 0.006$ \\
    \midrule
    \methodname\ (ours) & $\mathbf{0.064 \pm 0.000}$ \\ 
    \bottomrule
  \end{tabular}
  
\end{table}

\begin{table}[H]
  \centering
  \caption{RMSE calculated between the estimated and true marginal across different levels of label noise on the Colored MNIST dataset. $p$ defines label flip probability. Bold denotes best performance. \\}
  \label{table:noisy_cmnist}
  \small 
  \begin{tabular}{lcccc}
    \toprule
    & \multicolumn{4}{c}{Marginal RMSE $\downarrow$} \\
    \cmidrule(r){2-5}
    & $p=0.1$ & $p=0.2$ & $p=0.3$ & $p=0.4$\\
    \midrule
    CSVAE - No Adv. & $0.141 \pm 0.002$ & $0.141 \pm 0.003$ & $0.142 \pm 0.002$ & $0.143 \pm 0.002$ \\
    CSVAE & $0.135 \pm 0.022$ & $0.152 \pm 0.018$ & $0.181 \pm 0.007$ & $0.173 \pm 0.008$ \\
    HCSVAE - No Adv. & $0.150 \pm 0.001$ & $0.150 \pm 0.000$ & $0.151 \pm 0.000$ & $0.151 \pm 0.001$ \\
    HCSVAE & $0.139 \pm 0.003$ & $0.141 \pm 0.001$ & $0.141 \pm 0.001$ & $0.141 \pm 0.001$ \\
    DIVA & $0.115 \pm 0.011$ & $0.102 \pm 0.013$ & $0.106 \pm 0.010$ & $0.113 \pm 0.014$ \\
    CCVAE & $0.092 \pm 0.002$ & $0.103 \pm 0.014$ & $0.099 \pm 0.011$ & $0.092 \pm 0.005$ \\
    \midrule
    \methodname\ (ours) & $\mathbf{0.073 \pm 0.001}$ & $\mathbf{0.083 \pm 0.004}$ & $\mathbf{0.087 \pm 0.002}$ &  $\mathbf{0.087 \pm 0.001}$ \\ 
    \bottomrule
  \end{tabular}
\end{table}

\begin{table}[H]
  \caption{Model performances on the CelebA-Glasses dataset. Bold denotes best performance.}
  \label{suptable:celeba-res}
  \centering
  \begin{tabular}{lccc}
    \\ \toprule
    & $I(z;w) \downarrow$ & NLL ($\downarrow$) \\
    \midrule
    CSVAE - No Adv. & $0.048 \pm 0.014$ & $137.522 \pm 0.155$ \\
    CSVAE & $0.079 \pm 0.029$ & $145.989 \pm 0.336$ \\
    HCSVAE - No Adv. & $0.055 \pm 0.012$ & $131.813 \pm 0.21$ \\ 
    HCSVAE & $0.055 \pm 0.014$ & $137.319 \pm 0.265$ \\
    DIVA & $0.188 \pm  0.028$ & $143.528 \pm 0.02$  \\
    CCVAE & $0.083 \pm 0.022$ & $\mathbf{131.764 \pm 0.006}$  \\

    \midrule
    \methodname\ (ours) & $\mathbf{0.030 \pm 0.011}$ & $135.677 \pm 0.007$ \\ 
    \methodname\ - Common (ours) & --- & $374.114 \pm 0.05$ \\ 

    \bottomrule
  \end{tabular}
\end{table}

%%%%%%%%%%%%%%%%%%%%%%%%%%%%%%%%%%%%%%%%%%%%%

\section{Additional disentanglement metrics} \label{sec:metrics}

We provide an extended disentanglement assessment using multiple metrics. Because mutual information is difficult to estimate reliably, we report two estimators—MINE and kNN. Although their absolute values differ, the relative rankings of the methods remain consistent as can be seen in the ranking tables. In addition to these mutual-information estimates, we also report the following metrics, which quantify the level of label information captured by $w$ compared to $z$ :

\paragraph{Mutual Information Gap (MIG) } 
\begin{equation*}
    \operatorname{MIG}(w;z) = \frac{I(y;w) - I(y;z)}{H(y)}
\end{equation*}

\paragraph{Mutual Information Completeness (MIC)} 
\begin{equation*}
    \operatorname{MIC}(w;z) = \frac{I(y;w)}{I(y;w) + I(y;z)}
\end{equation*}

\subsection{Parametric model}
CSVAE and its variants impose a fully separable prior, thereby forcing separability even when the true latent structure is not separable (see Table \ref{table:1d_example}). In contrast, \methodname{} learns informative conditional embeddings that closely track the true posterior without requiring ground-truth knowledge of a truncated or fully separable prior, and it outperforms both DIVA and CCVAE.

Replacing the prior in \methodname{} with a fully separable predefined prior on $w$ yields consistent embeddings  with the ground-truth structure while retaining the benefits of separability.

\begin{table}[H]
    \centering
    \caption{Additional disentanglement metrics calculated with kNN mutual information estimation for the parametric model dataset with $k=20$. Bold indicates closest to true posterior within group.}
    \scalebox{0.75}{
    \begin{tabular}{clcccccc}
    \\ \toprule
    Assumption & Model & $I(y; z)$ & $I(y; w)$ & $I(w; z)$ & $\operatorname{MIG}(w;z)$ & $\operatorname{MIC}(w;z)$ & $I(w;z \mid y)$\\
    \midrule
    \multirow{4}{*}{Fully Separable} 
      & CSVAE - No Adv. &  $ \mathbf{0.069 \pm  0.034}$ & $ 0.634 \pm  0.002$ & $ \mathbf{0.098 \pm  0.034}$ & $ 0.063 \pm  0.004$ & $ \mathbf{0.904 \pm  0.048}$ & $0.000 \pm  0.002$\\
      & CSVAE        &  $ 0.024 \pm  0.048$ & $ 0.620 \pm  0.044$ & $ 0.047 \pm  0.050$ & $ \mathbf{0.067 \pm  0.010}$ & $ 0.963 \pm  0.073$ & $ 0.013 \pm  0.009$ \\
      & HCSVAE - No Adv.&  $ 0.000 \pm  0.000$ & $ \mathbf{0.643 \pm  0.000}$ & $ 0.000 \pm  0.001$ & $ 0.072 \pm  0.000$ & $ 1.000 \pm  0.000$ & $ 0.000 \pm  0.000$\\
      & HCSVAE       &  $ 0.000 \pm  0.000$ & $ \mathbf{0.643 \pm  0.001}$ & $ 0.001 \pm  0.001$ & $ 0.072 \pm  0.000$ & $ 1.000 \pm  0.000$ & $ 0.000 \pm  0.001$\\
      & DisCoVR (CSVAE prior) & $ 0.000 \pm  0.000$ & $ \mathbf{0.643 \pm  0.000}$ & $ 0.051 \pm  0.007$ & $ 0.072 \pm  0.000$ & $ 1.000 \pm  0.000$ & $ \mathbf{0.031 \pm  0.005}$ \\
    \midrule
    \multirow{3}{*}{Flexible}
      & DIVA         &  $ 0.021 \pm  0.042$ & $ 0.091 \pm  0.046$ & $ 0.000 \pm  0.000$ & $ 0.008 \pm  0.010$ & $ 0.800 \pm  0.400$ & $ 0.000 \pm  0.000$ \\
      & CCVAE        &  $ \mathbf{0.022 \pm  0.043}$ & $ 0.090 \pm  0.045$ & $ 0.000 \pm  0.000$ & $ 0.008 \pm  0.010$ & $ 0.800 \pm  0.400$ & $ 0.000 \pm  0.000$\\
      & DisCoVR (our prior) &  $ 0.010 \pm  0.006$ & $ \mathbf{0.151 \pm  0.007}$ & $ \mathbf{0.108 \pm  0.029}$ & $ \mathbf{0.016 \pm  0.001}$ & $ \mathbf{0.938 \pm  0.035}$ & $ \mathbf{0.072 \pm  0.020}$ \\
    \midrule
    \multirow{2}{*}{Fully Separable}
      & Posterior (no truncation) &  $ 0.057 \pm  0.001$ & $ 0.057 \pm  0.000$ & $ 0.144 \pm  0.003$ & $0.000 \pm  0.000$ & $ 0.499 \pm  0.006$ & $ 0.090 \pm  0.002$ \\
      & True Posterior            &  $ 0.058 \pm  0.003$ & $ 0.643 \pm  0.000$ & $ 0.144 \pm  0.005$ & $ 0.066 \pm  0.000$ & $ 0.917 \pm  0.004$ & $ 0.055 \pm  0.003$\\
    \bottomrule
    \end{tabular}}
\end{table}

\begin{table}[H]
    \centering
    \caption{Additional disentanglement metrics calculated with MINE mutual information estimation for the parametric model dataset. Bold indicates closest to true posterior within group.}
    \scalebox{0.75}{
    \begin{tabular}{clcccccc}
    \\ \toprule
    Assumption & Model & $I(y; z)$ & $I(y; w)$ & $I(w; z)$ & $\operatorname{MIG}(w;z)$ & $\operatorname{MIC}(w;z)$ & $I(w;z \mid y)$\\
    \midrule
    \multirow{4}{*}{Fully Separable}
      & CSVAE - No Adv.      & $ \mathbf{0.096 \pm  0.037}$ & $ 0.528 \pm  0.026$ & $ \mathbf{0.096 \pm  0.034}$ & $ \mathbf{0.048 \pm  0.006}$ & $ \mathbf{0.848 \pm  0.057}$ & $ 0.001 \pm  0.001$\\
      & CSVAE             & $ 0.033 \pm  0.055$ & $ \mathbf{0.526 \pm  0.045}$ & $ 0.031 \pm  0.045$ & $ 0.055 \pm  0.010$ & $ 0.945 \pm  0.091$ & $ 0.009 \pm  0.005$ \\
      & HCSVAE - No Adv.     & $ 0.000 \pm  0.000$ & $ 0.543 \pm  0.018$ & $ 0.000 \pm  0.000$ & $ 0.061 \pm  0.002$ & $ 1.000 \pm  0.000$ & $ 0.000 \pm  0.000$ \\
      & HCSVAE            & $ 0.000 \pm  0.000$ & $ 0.543 \pm  0.022$ & $ 0.000 \pm  0.000$ & $ 0.061 \pm  0.002$ & $ 1.000 \pm  0.000$ & $ 0.001 \pm  0.000$\\
      & DisCoVR (CSVAE prior) & $ 0.020 \pm  0.004$ & $ 0.543 \pm  0.018$ & $ 0.033 \pm  0.005$ & $ 0.058 \pm  0.002$ & $ 0.964 \pm  0.008$ & $ \mathbf{0.030 \pm  0.004}$ \\
    \midrule
    \multirow{3}{*}{Flexible}
      & DIVA              & $ 0.027 \pm  0.053$ & $ 0.113 \pm  0.056$ & $ 0.001 \pm  0.001$ & $ 0.010 \pm  0.012$ & $ 0.798 \pm  0.398$ & $ 0.001 \pm  0.000$ \\
      & CCVAE             & $ 0.026 \pm  0.052$ & $ 0.115 \pm  0.058$ & $ 0.001 \pm  0.001$ & $ 0.010 \pm  0.012$ & $ 0.799 \pm  0.399$ & $ 0.001 \pm  0.000$\\
      & DisCoVR (ours)    & $ \mathbf{0.037 \pm  0.006}$ & $ \mathbf{0.176 \pm  0.008}$ & $ \mathbf{0.109 \pm  0.025}$ & $ \mathbf{0.016 \pm  0.001}$ & $ \mathbf{0.825 \pm  0.026}$ & $ \mathbf{0.073 \pm  0.018}$\\
    \midrule
    \multirow{2}{*}{Fully Separable}
      & Posterior (no truncation) &  $ 0.084 \pm  0.003$ & $ 0.083 \pm  0.003$ & $ 0.137 \pm  0.004$ & $0.000 \pm  0.000$ & $ 0.497 \pm  0.009$ & $ 0.088 \pm  0.003$\\
      & True Posterior            & $ 0.085 \pm  0.005$ & $ 0.493 \pm  0.011$ & $ 0.139 \pm  0.005$ & $ 0.046 \pm  0.002$ & $ 0.853 \pm  0.009$ & $ 0.057 \pm  0.004$\\
    \bottomrule
    \end{tabular}}
\end{table}

\begin{table}[H]
    \centering
    \caption{Rank (1 = closest to True Posterior) of each method with respect to the true posterior for metrics calculated with kNN mutual information estimation with $k=20$. Colors indicate rank within each block: red = worse (farther), green = better (closer).}
    \scalebox{0.75}{
    \begin{tabular}{clcccccc}
    \\ \toprule
    Assumption & Model 
      & $I(y; z)$ 
      & $I(y; w)$ 
      & $I(w; z)$ 
      & $\operatorname{MIG}(w;z)$ 
      & $\operatorname{MIC}(w;z)$ 
      & $I(w;z \mid y)$ \\
    \midrule
    
    & CSVAE - No Adv. 
        & \cellcolor{Upper!85!Lower} 1 
        & \cellcolor{Upper!40!Lower} 4 
        & \cellcolor{Upper!85!Lower} 1 
        & \cellcolor{Upper!70!Lower} 2 
        & \cellcolor{Upper!85!Lower} 1 
        & \cellcolor{Upper!55!Lower} 3 \\
    & CSVAE 
        & \cellcolor{Upper!70!Lower} 2 
        & \cellcolor{Upper!25!Lower} 5 
        & \cellcolor{Upper!55!Lower} 3 
        & \cellcolor{Upper!85!Lower} 1 
        & \cellcolor{Upper!70!Lower} 2 
        & \cellcolor{Upper!70!Lower} 2 \\
    & HCSVAE - No Adv. 
        & \cellcolor{Upper!55!Lower} 3 
        & \cellcolor{Upper!85!Lower} 1 
        & \cellcolor{Upper!25!Lower} 5 
        & \cellcolor{Upper!55!Lower} 3 
        & \cellcolor{Upper!55!Lower} 3 
        & \cellcolor{Upper!55!Lower} 3 \\
    & HCSVAE 
        & \cellcolor{Upper!55!Lower} 3 
        & \cellcolor{Upper!85!Lower} 1 
        & \cellcolor{Upper!40!Lower} 4 
        & \cellcolor{Upper!55!Lower} 3 
        & \cellcolor{Upper!55!Lower} 3 
        & \cellcolor{Upper!55!Lower} 3 \\
    \multirow{-5}{*}{Fully Separable} & DisCoVR (CSVAE prior) 
        & \cellcolor{Upper!55!Lower} 3 
        & \cellcolor{Upper!85!Lower} 1 
        & \cellcolor{Upper!70!Lower} 2 
        & \cellcolor{Upper!55!Lower} 3 
        & \cellcolor{Upper!55!Lower} 3 
        & \cellcolor{Upper!85!Lower} 1 \\
    \midrule
    & DIVA 
        & \cellcolor{Upper!70!Lower} 2 
        & \cellcolor{Upper!70!Lower} 2 
        & \cellcolor{Upper!70!Lower} 2 
        & \cellcolor{Upper!70!Lower} 2 
        & \cellcolor{Upper!70!Lower} 2 
        & \cellcolor{Upper!70!Lower} 2 \\
    & CCVAE 
        & \cellcolor{Upper!85!Lower} 1 
        & \cellcolor{Upper!55!Lower} 3 
        & \cellcolor{Upper!70!Lower} 2 
        & \cellcolor{Upper!70!Lower} 2 
        & \cellcolor{Upper!70!Lower} 2 
        & \cellcolor{Upper!70!Lower} 2 \\
    \multirow{-3}{*}{ Flexible} & DisCoVR (our prior) 
        & \cellcolor{Upper!55!Lower} 3 
        & \cellcolor{Upper!85!Lower} 1 
        & \cellcolor{Upper!85!Lower} 1 
        & \cellcolor{Upper!85!Lower} 1 
        & \cellcolor{Upper!85!Lower} 1 
        & \cellcolor{Upper!85!Lower} 1 \\
    \bottomrule
    \end{tabular}}
\end{table}

\begin{table}[H]
    \centering
    \caption{Rank (1 = closest to True Posterior) of each method with respect to the True Posterior for metrics calculated with MINE mutual information estimation. Colors indicate rank within each block: red = worse (farther), green = better (closer).}
    \scalebox{0.75}{
    \begin{tabular}{clcccccc}
    \\ \toprule
    Assumption & Model 
      & $I(y; z)$ 
      & $I(y; w)$ 
      & $I(w; z)$ 
      & $\operatorname{MIG}(w;z)$ 
      & $\operatorname{MIC}(w;z)$ 
      & $I(w;z \mid y)$ \\
    \midrule
    
       &  CSVAE - No Adv. 
        & \cellcolor{Upper!85!Lower} 1 
        & \cellcolor{Upper!70!Lower} 2 
        & \cellcolor{Upper!85!Lower} 1 
        & \cellcolor{Upper!85!Lower} 1 
        & \cellcolor{Upper!85!Lower} 1 
        & \cellcolor{Upper!55!Lower} 3 \\
       &  CSVAE 
        & \cellcolor{Upper!70!Lower} 2 
        & \cellcolor{Upper!85!Lower} 1 
        & \cellcolor{Upper!55!Lower} 3 
        & \cellcolor{Upper!70!Lower} 2 
        & \cellcolor{Upper!70!Lower} 2 
        & \cellcolor{Upper!70!Lower} 2 \\
       &  HCSVAE - No Adv. 
        & \cellcolor{Upper!40!Lower} 4 
        & \cellcolor{Upper!55!Lower} 3 
        & \cellcolor{Upper!40!Lower} 4 
        & \cellcolor{Upper!40!Lower} 4 
        & \cellcolor{Upper!40!Lower} 4 
        & \cellcolor{Upper!25!Lower} 5 \\
       &  HCSVAE 
        & \cellcolor{Upper!40!Lower} 4 
        & \cellcolor{Upper!55!Lower} 3 
        & \cellcolor{Upper!40!Lower} 4 
        & \cellcolor{Upper!40!Lower} 4 
        & \cellcolor{Upper!40!Lower} 4 
        & \cellcolor{Upper!55!Lower} 3 \\
      \multirow{-5}{*}{Fully Separable} &  DisCoVR (CSVAE prior) 
        & \cellcolor{Upper!55!Lower} 3 
        & \cellcolor{Upper!55!Lower} 3 
        & \cellcolor{Upper!70!Lower} 2 
        & \cellcolor{Upper!55!Lower} 3 
        & \cellcolor{Upper!55!Lower} 3 
        & \cellcolor{Upper!85!Lower} 1 \\
    \midrule
       &  DIVA 
        & \cellcolor{Upper!70!Lower} 2 
        & \cellcolor{Upper!55!Lower} 3 
        & \cellcolor{Upper!70!Lower} 2 
        & \cellcolor{Upper!70!Lower} 2 
        & \cellcolor{Upper!55!Lower} 3 
        & \cellcolor{Upper!70!Lower} 2 \\
       &  CCVAE 
        & \cellcolor{Upper!55!Lower} 3 
        & \cellcolor{Upper!70!Lower} 2 
        & \cellcolor{Upper!70!Lower} 2 
        & \cellcolor{Upper!70!Lower} 2 
        & \cellcolor{Upper!70!Lower} 2 
        & \cellcolor{Upper!70!Lower} 2 \\
      \multirow{-3}{*}{Flexible} &  DisCoVR (ours) 
        & \cellcolor{Upper!85!Lower} 1 
        & \cellcolor{Upper!85!Lower} 1 
        & \cellcolor{Upper!85!Lower} 1 
        & \cellcolor{Upper!85!Lower} 1 
        & \cellcolor{Upper!85!Lower} 1 
        & \cellcolor{Upper!85!Lower} 1 \\
    \bottomrule
    \end{tabular}}
\end{table} 

\subsection{Noisy Swiss Roll}

When the observed labels are noisy, \methodname{} outperforms other methods, obtaining embeddings close to the ground truth. 

\begin{table}[H]
    \centering
    \caption{Additional disentanglement metrics calculated with kNN mutual information estimation for the Noisy Swiss Roll ($p=0.3$) dataset with $k=20$. Bold indicates closest to ground truth within group.}
    \scalebox{0.75}{
    \begin{tabular}{clcccccc}
    \\ \toprule
    Assumption & Model & $I(y; z)$ & $I(y; w)$ & $I(w; z)$ & $\operatorname{MIG}(w;z)$ & $\operatorname{MIC}(w;z)$ & $I(w;z \mid y)$\\
    \midrule
    \multirow{4}{*}{Fully Separable}
      & CSVAE - No Adv. & $ 0.041 \pm  0.007$ & $ 0.525 \pm  0.221$ & $ 0.362 \pm  0.180$ & $ 0.057 \pm  0.026$ & $ 0.888 \pm  0.098$ & $ 0.266 \pm  0.152$\\
      & CSVAE        & $ 0.018 \pm  0.026$ & $ \mathbf{0.429 \pm  0.254}$ & $ 0.240 \pm  0.181$ & $ \mathbf{0.048 \pm  0.032}$ & $ 0.912 \pm  0.129$ & $ 0.186 \pm  0.146$ \\
      & HCSVAE - No Adv.& $ 0.029 \pm  0.007$ & $ 0.642 \pm  0.000$ & $ 0.065 \pm  0.013$ & $ 0.072 \pm  0.001$ & $ 0.957 \pm  0.010$ & $ 0.009 \pm  0.019$  \\
      & HCSVAE       & $ \mathbf{0.001 \pm  0.002}$ & $ 0.641 \pm  0.001$ & $ \mathbf{0.005 \pm  0.004}$ & $ 0.075 \pm  0.000$ & $ \mathbf{0.999 \pm  0.003}$ & $ \mathbf{0.000 \pm  0.000}$ \\
    \midrule
    \multirow{3}{*}{Flexible}
      & DIVA         & $ 0.034 \pm  0.013$ & $ 0.036 \pm  0.011$ & $ 2.633 \pm  0.360$ & $ 0.000 \pm  0.003$ & $ 0.515 \pm  0.159$ & $ 2.185 \pm  0.332$ \\
      & CCVAE        & $ 0.040 \pm  0.015$ & $ 0.030 \pm  0.007$ & $ 2.952 \pm  0.124$ & $-0.001 \pm  0.003$ & $ 0.447 \pm  0.153$ & $ 2.462 \pm  0.118$ \\
      & DisCoVR (ours) & $ \mathbf{0.000 \pm  0.000}$ & $ \mathbf{0.049 \pm  0.002}$ & $ \mathbf{0.029 \pm  0.011}$ & $ \mathbf{0.006 \pm  0.000}$ & $ \mathbf{1.000 \pm  0.000}$ & $ \mathbf{0.014 \pm  0.008}$ \\
    \midrule
    \multirow{1}{*}{Noisy}
      & Ground Truth &  $ 0.000 \pm  0.000$ & $ 0.055 \pm  0.002$ & $ 0.000 \pm  0.000$ & $ 0.007 \pm  0.000$ & $ 1.000 \pm  0.000$ & $ 0.000 \pm  0.000$ \\
    \bottomrule
    \end{tabular}}
\end{table}

\begin{table}[H]
    \centering
    \caption{Additional disentanglement metrics calculated with MINE mutual information estimation for the Noisy Swiss Roll ($p=0.3$) dataset. Bold indicates closest to ground truth within group.}
    \scalebox{0.75}{
    \begin{tabular}{clcccccc}
    \\ \toprule
    Assumption & Model & $I(y; z)$ & $I(y; w)$ & $I(w; z)$ & $\operatorname{MIG}(w;z)$ & $\operatorname{MIC}(w;z)$ & $I(w;z \mid y)$\\
    \midrule
    & CSVAE - No Adv.& $ 0.046 \pm  0.022$ & $ 0.422 \pm  0.186$ & $ 0.050 \pm  0.020$ & $ 0.044 \pm  0.023$ & $ 0.834 \pm  0.184$ & $ 0.029 \pm  0.017$ \\
    & CSVAE &  $ 0.023 \pm  0.027$ & $ \mathbf{0.373 \pm  0.232}$ & $ 0.027 \pm  0.020$ & $ \mathbf{0.041 \pm  0.028}$ & $ 0.877 \pm  0.198$ & $ 0.024 \pm  0.017$ \\
    & HCSVAE - No Adv. & $ 0.023 \pm  0.014$ & $ 0.585 \pm  0.011$ & $ 0.026 \pm  0.012$ & $ 0.066 \pm  0.002$ & $ 0.963 \pm  0.021$ & $ 0.006 \pm  0.002$\\
    \multirow{-4}{*}{Fully Separable} & HCSVAE & $ \mathbf{0.002 \pm  0.000}$ & $ 0.570 \pm  0.011$ & $ \mathbf{0.002 \pm  0.001}$ & $ 0.067 \pm  0.001$ & $ \mathbf{0.997 \pm  0.001}$ & $ \mathbf{0.003 \pm  0.001}$ \\
    \midrule
    & DIVA & $ 0.041 \pm  0.024$ & $ 0.043 \pm  0.026$ & $ 0.313 \pm  0.084$ & $ \mathbf{0.000 \pm  0.006}$ & $ 0.507 \pm  0.296$ & $ 0.345 \pm  0.065$ \\
    & CCVAE & $ 0.056 \pm  0.020$ & $\mathbf{0.036 \pm  0.020}$ & $ 0.507 \pm  0.114$ & $-0.002 \pm  0.004$ & $ 0.390 \pm  0.226$ & $ 0.494 \pm  0.099$\\
    \multirow{-3}{*}{Flexible} & DisCoVR (ours) & $ \mathbf{0.001 \pm  0.000}$ & $ 0.069 \pm  0.002$ & $ \mathbf{0.004 \pm  0.002}$ & $ 0.008 \pm  0.000$ & $ \mathbf{0.983 \pm  0.004}$ & $ \mathbf{0.006 \pm  0.002}$\\
    \midrule
    \multirow{1}{*}{Noisy} & Ground Truth & $0.000 \pm  0.000$ & $ 0.024 \pm  0.018$ & $ 0.000 \pm  0.001$ & $ 0.003 \pm  0.002$ & $ 0.985 \pm  0.048$ & $ 0.002 \pm  0.001$\\
    \bottomrule
    \end{tabular}}
\end{table}

\begin{table}[H]
    \centering
    \caption{Rank (1 = closest to Ground Truth) of each method with respect to the Ground Truth for metrics calculated with kNN mutual information estimation with $k=20$ on the Noisy Swiss Roll ($p=0.3$) dataset. Colors indicate rank within each block: red = worse (farther), green = better (closer).}
    \scalebox{0.75}{
    \begin{tabular}{clcccccc}
    \\ \toprule
    Assumption & Method 
      & $I(y; z)$ 
      & $I(y; w)$ 
      & $I(w; z)$ 
      & $\operatorname{MIG}(w;z)$ 
      & $\operatorname{MIC}(w;z)$ 
      & $I(w;z \mid y)$ \\
    \midrule
    
       & CSVAE - No Adv. 
        & \cellcolor{Upper!40!Lower} 4 
        & \cellcolor{Upper!70!Lower} 2 
        & \cellcolor{Upper!40!Lower} 4 
        & \cellcolor{Upper!70!Lower} 2 
        & \cellcolor{Upper!40!Lower} 4 
        & \cellcolor{Upper!40!Lower} 4 \\
       & CSVAE 
        & \cellcolor{Upper!70!Lower} 2 
        & \cellcolor{Upper!85!Lower} 1 
        & \cellcolor{Upper!55!Lower} 3 
        & \cellcolor{Upper!85!Lower} 1 
        & \cellcolor{Upper!55!Lower} 3 
        & \cellcolor{Upper!55!Lower} 3 \\
       & HCSVAE - No Adv. 
        & \cellcolor{Upper!55!Lower} 3 
        & \cellcolor{Upper!40!Lower} 4 
        & \cellcolor{Upper!70!Lower} 2 
        & \cellcolor{Upper!55!Lower} 3 
        & \cellcolor{Upper!70!Lower} 2 
        & \cellcolor{Upper!70!Lower} 2 \\
      \multirow{-4}{*}{Fully Separable} & HCSVAE 
        & \cellcolor{Upper!85!Lower} 1 
        & \cellcolor{Upper!55!Lower} 3 
        & \cellcolor{Upper!85!Lower} 1 
        & \cellcolor{Upper!40!Lower} 4 
        & \cellcolor{Upper!85!Lower} 1 
        & \cellcolor{Upper!85!Lower} 1 \\
    \midrule
       & DIVA 
        & \cellcolor{Upper!70!Lower} 2 
        & \cellcolor{Upper!70!Lower} 2 
        & \cellcolor{Upper!70!Lower} 2 
        & \cellcolor{Upper!70!Lower} 2 
        & \cellcolor{Upper!70!Lower} 2 
        & \cellcolor{Upper!70!Lower} 2 \\
       & CCVAE 
        & \cellcolor{Upper!55!Lower} 3 
        & \cellcolor{Upper!55!Lower} 3 
        & \cellcolor{Upper!55!Lower} 3 
        & \cellcolor{Upper!55!Lower} 3 
        & \cellcolor{Upper!55!Lower} 3 
        & \cellcolor{Upper!55!Lower} 3 \\
      \multirow{-3}{*}{ Flexible} & DisCoVR (ours) 
        & \cellcolor{Upper!85!Lower} 1 
        & \cellcolor{Upper!85!Lower} 1 
        & \cellcolor{Upper!85!Lower} 1 
        & \cellcolor{Upper!85!Lower} 1 
        & \cellcolor{Upper!85!Lower} 1 
        & \cellcolor{Upper!85!Lower} 1 \\
    \bottomrule
    \end{tabular}}
\end{table}

\begin{table}[H]
    \centering
    \caption{Rank (1 = closest to Ground Truth) of each method with respect to the Ground Truth for metrics calculated with MINE mutual information estimation on the Noisy Swiss Roll ($p=0.3$) dataset. Colors indicate rank within each block: red = worse (farther), green = better (closer).}
    \scalebox{0.75}{
    \begin{tabular}{clcccccc}
    \\ \toprule
    Assumption & Model
      & $I(y; z)$ 
      & $I(y; w)$ 
      & $I(w; z)$ 
      & $\operatorname{MIG}(w;z)$ 
      & $\operatorname{MIC}(w;z)$ 
      & $I(w;z \mid y)$ \\
    \midrule
    
       & CSVAE - No Adv. 
        & \cellcolor{Upper!40!Lower} 4 
        & \cellcolor{Upper!70!Lower} 2 
        & \cellcolor{Upper!40!Lower} 4 
        & \cellcolor{Upper!70!Lower} 2 
        & \cellcolor{Upper!40!Lower} 4 
        & \cellcolor{Upper!40!Lower} 4 \\
       & CSVAE 
        & \cellcolor{Upper!70!Lower} 2 
        & \cellcolor{Upper!85!Lower} 1 
        & \cellcolor{Upper!55!Lower} 3 
        & \cellcolor{Upper!85!Lower} 1 
        & \cellcolor{Upper!55!Lower} 3 
        & \cellcolor{Upper!55!Lower} 3 \\
       & HCSVAE - No Adv. 
        & \cellcolor{Upper!70!Lower} 2 
        & \cellcolor{Upper!40!Lower} 4 
        & \cellcolor{Upper!70!Lower} 2 
        & \cellcolor{Upper!55!Lower} 3 
        & \cellcolor{Upper!70!Lower} 2 
        & \cellcolor{Upper!70!Lower} 2 \\
      \multirow{-4}{*}{Fully Separable} & HCSVAE 
        & \cellcolor{Upper!85!Lower} 1 
        & \cellcolor{Upper!55!Lower} 3 
        & \cellcolor{Upper!85!Lower} 1 
        & \cellcolor{Upper!40!Lower} 4 
        & \cellcolor{Upper!85!Lower} 1 
        & \cellcolor{Upper!85!Lower} 1 \\
    \midrule
       & DIVA 
        & \cellcolor{Upper!70!Lower} 2 
        & \cellcolor{Upper!70!Lower} 2 
        & \cellcolor{Upper!70!Lower} 2 
        & \cellcolor{Upper!85!Lower} 1 
        & \cellcolor{Upper!70!Lower} 2 
        & \cellcolor{Upper!70!Lower} 2 \\
       & CCVAE 
        & \cellcolor{Upper!55!Lower} 3 
        & \cellcolor{Upper!85!Lower} 1 
        & \cellcolor{Upper!55!Lower} 3 
        & \cellcolor{Upper!70!Lower} 2 
        & \cellcolor{Upper!55!Lower} 3 
        & \cellcolor{Upper!55!Lower} 3 \\
      \multirow{-3}{*}{Flexible} & DisCoVR (ours) 
        & \cellcolor{Upper!85!Lower} 1 
        & \cellcolor{Upper!55!Lower} 3 
        & \cellcolor{Upper!85!Lower} 1 
        & \cellcolor{Upper!70!Lower} 2 
        & \cellcolor{Upper!85!Lower} 1 
        & \cellcolor{Upper!85!Lower} 1 \\
    \bottomrule
    \end{tabular}}
\end{table}

\section{Ablations on model components} \label{sec:ablations}
Here we evaluate two variations of our model:
(1) applying the classifier directly to $z$, and (2) replacing the conditional prior on $w$ with a standard Gaussian.

When training the classifier directly on $z$ we were able to achieve results qualitatively similar to those obtained using the reconstruction  $\hat{x}$, but doing so requires substantially more parameter tuning.

An unconditional standard Gaussian prior for $w$ causes $w$ to collapse into a representation redundant with $z$, removing meaningful separation.

\begin{figure}[H]
    \centering
    \includegraphics[width=\linewidth]{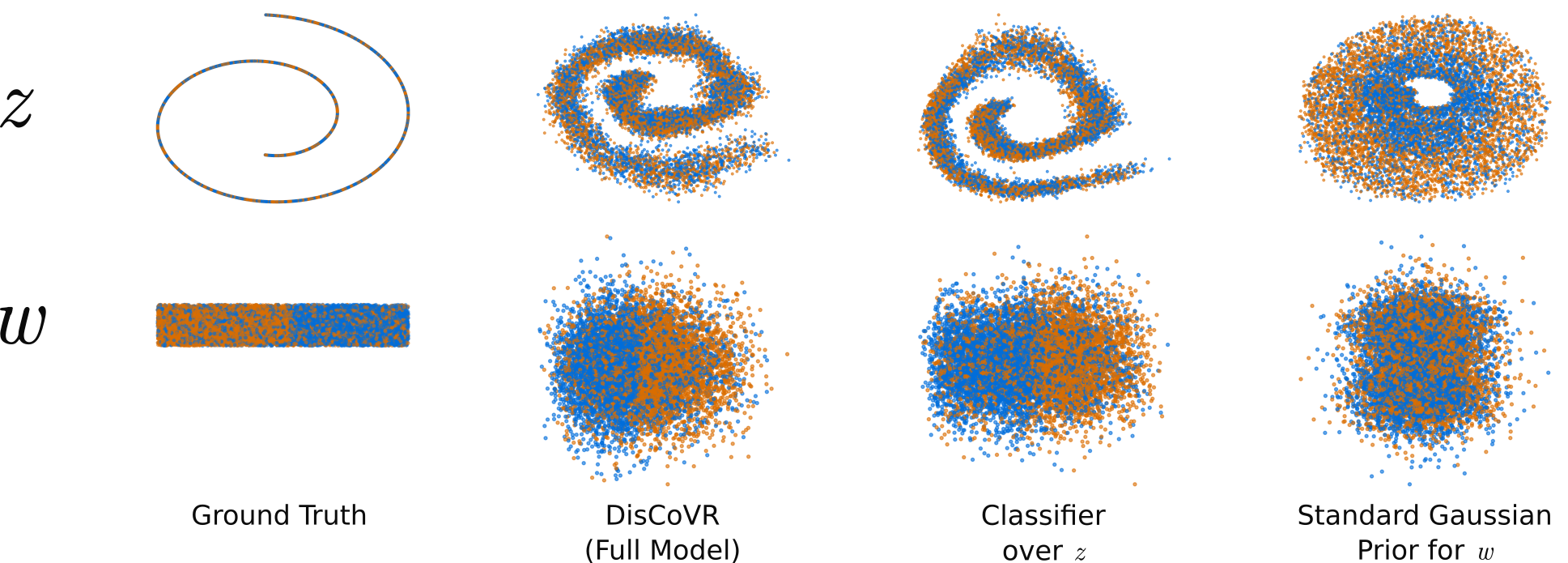}
    \vspace{0.3em}
    \caption{Ablation study on the Noisy Swiss Roll ($p=0.3$) dataset.}
    \label{fig:ablations_nsr}
\end{figure}

\begin{figure}[H]
    \centering
    \includegraphics[width=\linewidth]{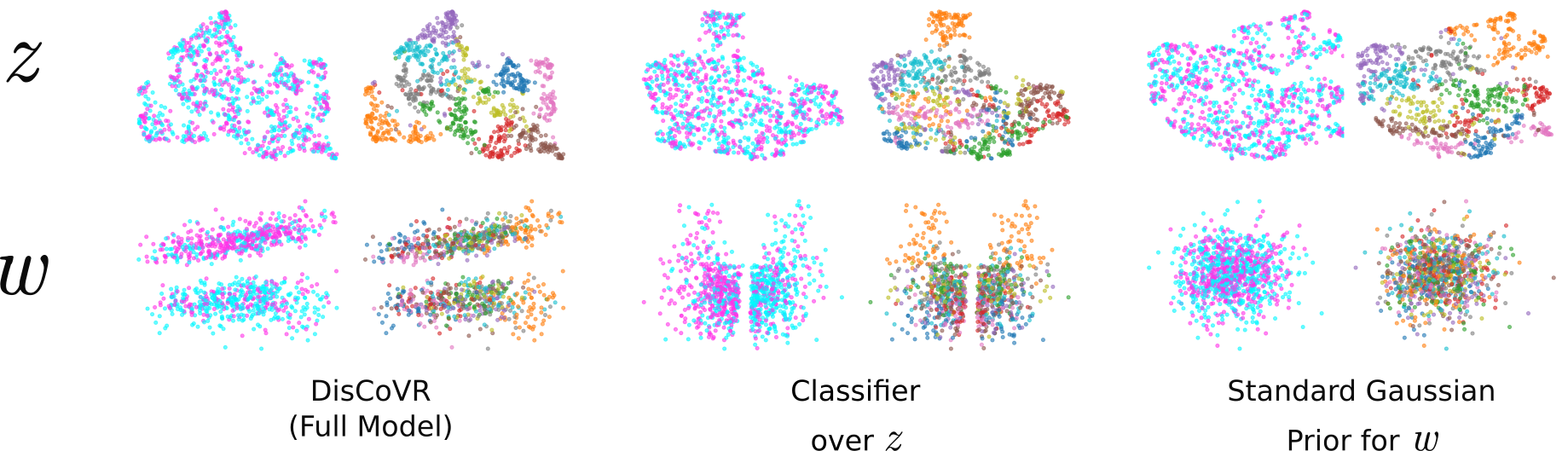}
    \vspace{0.3em}
    \caption{Ablation study on the Noisy Colored MNIST ($p=0.3$) dataset. For each setting: left column denotes coloring by noisy labels, right column denotes coloring by digit (shape, not included in the label).}
    \label{fig:ablations_cmnist}
\end{figure}

We consider an additional ablation of the adversarial component, by varying the weight of the adversarial loss on the Noisy Swiss Roll dataset. The results for kNN MI estimation and MINE MI estimation are presented in Tables \ref{table:knn_adv_abl} and \ref{table:mine_adv_abl} respectively.

\begin{table}[H]
  \centering
  \caption{Information metrics estimated using kNN mutual information across different adversarial weights for the Noisy Swiss Roll dataset. Arrows indicate the desired direction. Bold denotes best performance.}
  \label{table:knn_adv_abl}
  \small
  \begin{tabular}{lcccccc}
    \toprule
    & $I(y;z)\downarrow$ 
    & $I(y;w)\uparrow$ 
    & $I(w;z)\downarrow$ 
    & $\mathrm{MIG}(w;z)\uparrow$ 
    & $\mathrm{MIC}(w;z)\uparrow$ 
    & $I(w;z \mid y)\downarrow$ \\
    \midrule
    Adv. $=0$ & $0.049$ & $0.000$ & $0.887$ & $-0.006$ & $0.000$ & $0.812$ \\
    Adv. $=2$ & $0.051$ & $0.000$ & $0.489$ & $-0.006$ & $0.000$ & $0.427$ \\
    Adv. $=4$ & $0.052$ & $0.000$ & $0.480$ & $-0.006$ & $0.000$ & $0.419$ \\
    Adv. $=6$ & $0.007$ & $0.045$ & $0.179$ & $0.005$ & $0.873$ & $0.112$ \\
    Adv. $=8^\dagger$ 
    & $\mathbf{0.000}$ 
    & $\mathbf{0.050}$ 
    & $\mathbf{0.032}$ 
    & $\mathbf{0.006}$ 
    & $\mathbf{1.000}$ 
    & $\mathbf{0.017}$ \\
    \bottomrule
  \end{tabular}

  \vspace{0.5em}
  \footnotesize{$^\dagger$ The value used in the paper.}
\end{table}

\begin{table}[H]
  \centering
  \caption{Information metrics estimated using MINE mutual information across different adversarial weights for the Noisy Swiss Roll dataset. Arrows indicate the desired direction. Bold denotes best performance.}
  \label{table:mine_adv_abl}
  \small
  \begin{tabular}{lcccccc}
    \toprule
    & $I(y;z)\downarrow$ 
    & $I(y;w)\uparrow$ 
    & $I(w;z)\downarrow$ 
    & $\mathrm{MIG}(w;z)\uparrow$ 
    & $\mathrm{MIC}(w;z)\uparrow$ 
    & $I(w;z \mid y)\downarrow$ \\
    \midrule
    Adv. $=0$ & $0.067$ & $0.011$ & $0.222$ & $-0.007$ & $0.136$ & $0.249$ \\
    Adv. $=2$ & $0.051$ & $0.011$ & $0.105$ & $-0.005$ & $0.182$ & $0.118$ \\
    Adv. $=4$ & $0.051$ & $0.011$ & $0.101$ & $-0.005$ & $0.184$ & $0.111$ \\
    Adv. $=6$ & $0.003$ & $0.064$ & $0.015$ & $0.007$ & $0.950$ & $0.020$ \\
    Adv. $=8^\dagger$ 
    & $\mathbf{0.001}$ 
    & $\mathbf{0.066}$ 
    & $\mathbf{0.004}$ 
    & $\mathbf{0.008}$ 
    & $\mathbf{0.980}$ 
    & $\mathbf{0.008}$ \\
    \bottomrule
  \end{tabular}

  \vspace{0.5em}
  \footnotesize{$^\dagger$ The value used in the paper.}
\end{table}

\section{Additional experiment on CelebA-Hats} \label{sup:hats}
 We performed an additional experiment on the CelebA dataset, with the attribute $W\!earing\_hat$ denoting the $y$ label. Supplementary Table \ref{suptable:celeba-hats} outlines the results of this experiment. \methodname{} is the only method that exhibits high disentanglement for $z, w$ without compromising reconstruction quality.

\begin{table}[H]
  \caption{Model performances of a single experiment on CelebA-Hats. Bold denotes best performance.}
  \label{suptable:celeba-hats}
  \centering
  \begin{tabular}{lccc}
    \\ \toprule
    & $I(z;w) \downarrow$ & NLL ($\downarrow$) \\
    \midrule
    CSVAE - No Adv.  & 0.360 & 653.537 \\
    CSVAE  & 0.213 & \textbf{351.082} \\
    HCSVAE - No Adv.  & 0.135 & 2608.442 \\
    HCSVAE  & 0.192 & 673.674 \\
    DIVA  & 0.553 & 356.090 \\
    CCVAE  & 0.856 & 347.940 \\
    \midrule 
    \methodname\ (ours) & \textbf{0.059} & 353.271 \\
    \methodname\ (ours) - Common & - & 437.144 \\
    
    \bottomrule
  \end{tabular}
\end{table}

%%%%%%%%%%%%%%%%%%%%%%%%%%%%%%%%%%%%%%%%%%%%%%
\section{Implementation details} \label{sup:implement}
\subsection{Considerations and reproducibility}
We run all experiments on a single H100 GPU. Reported means and standard deviations for tables are conducted over 10 repetitions of the experiment with different random seeds. All models are trained using the AdamW \citep{loshchilov2018decoupled} optimizer until validation loss stops decreasing for 50 epochs. Wherever provided, we use mutual information neural estimation (MINE, \citet{pmlr-v80-belghazi18a}) and k-Nearest Neighbor (kNN) mutual information estimation \cite{Kraskov2004} to obtain mutual information estimates. For Naive Bayes classifiers, we use the implementation provided by \mbox{\emph{scikit-learn}} \citep{scikit-learn}. To use ideal hyperparameters for each method, we consult the original implementation whenever possible, and conduct a simple grid-search to produce originally described model behavior. 
Implementations of all methods compared in this study, including \methodname{}, as well as code to reproduce our results, are available at \url{https://github.com/Computational-Morphogenomics-Group/DISCoVeR}. 
Models compared in the study admit a weighting term for each term in the loss function, of which most are shared across different approaches. We use the following shorthands for each of the terms:
\begin{align*}
& \text{Rec.} \rightarrow \E_{q_{z \vert x}}\left[ \E_{q_{w \vert x,y}} [\log p(x\mid z,w)] \right] \\ 
& \KL(Z) \rightarrow \KL(q_{z\vert x}\,\Vert\,p_z) \\ 
& \KL(W) \rightarrow \KL\left(q_{w \vert x,y}\,\Vert\,p_{w\mid y}\right)\\
& \text{Adv.} \rightarrow -\E_{q_{z\vert x}}[\log g(y\mid z)] \\
& \text{Class.} \rightarrow \E_{q_{w\vert x,y}}[\log q(y\mid w)]\\
& \text{Rec. - ($Z$)} \rightarrow  \E_{q_{z \vert x}} [\log p(x\mid z)] 
\end{align*}
Below, we provide additional details for the hyperparameters used in each experiment, and any other external resources used to obtain the corresponding sections' results. In addition, we include details regarding runtime and memory footprint of running experiments with the models included in our study.

\begin{table}[H]
  \caption{Time spent per epoch during training for each dataset.}
  \label{suptable:time-epoch}
  \centering
  \begin{tabular}{lccccc}
    \\ \toprule
    & P.M. & N.S.R & CMNIST & CelebA & scRNA-seq \\
    \midrule
    CSVAE - No Adv. & 10.91s & 8.02s & 14s & 54.43s & 4.78s \\ 
    CSVAE &	12.71s & 8.98s & 14.41s & 74.68s & 4.99s \\ 
    HCSVAE - No Adv. & 15.6s & 10.92s & 17s & 44.3s & 6.34s \\ 
    HCSVAE & 16.51s  & 11.8s  & 16.8s  & 68.53s  & 5.3s \\ 
    DIVA & 11.1s  & 8s & 18.59s  & 48.96s  & 3.55s \\ 
    CCVAE & 12.1s  & 8.44s  & 17.9s  & 49.65s  & 5.25s \\ 
    \methodname (ours) & 12.18s  & 8.73s  & 21.8s  & 109.86s  & 6.09s \\ 
    \bottomrule
  \end{tabular}
\end{table}

\begin{table}[H]
  \caption{Model inference time for a single batch for each dataset.}
  \label{suptable:time-inf}
  \centering
  \begin{tabular}{lccccc}
    \\ \toprule
    & P.M. & N.S.R & CMNIST & CelebA & scRNA-seq \\
    \midrule
   CSVAE - No Adv.  & 72ms  & 28ms  & 57ms  & 329ms  & 100ms\\
    CSVAE  & 40ms  & 29ms  & 59ms  & 187ms  & 96ms\\
    HCSVAE - No Adv.  & 36ms  & 30ms  & 55ms  & 214ms  & 95ms\\
    HCSVAE  & 37ms  & 39ms  & 52ms  & 191ms  & 119ms\\
    DIVA  & 25ms  & 26ms  & 60ms  & 154ms  & 42ms\\
    CCVAE  & 27ms  & 28ms  & 42ms  & 163ms  & 93ms\\
    \methodname (ours)  & 32ms  & 27ms  & 63ms  & 205ms  & 123ms \\
    \bottomrule
  \end{tabular}
\end{table}

\begin{table}[H]
  \caption{Memory footprint of running an experiment for each dataset.}
  \label{suptable:mem-exp}
  \centering
  \begin{tabular}{lccccc}
    \\ \toprule
    & P.M. & N.S.R & CMNIST & CelebA & scRNA-seq \\
    \midrule
    CSVAE - No Adv. & 53 MiB  & 255 MiB  & 1988 MiB  & 4868 MiB  & 298 MiB \\
    CSVAE  & 253 MiB  & 255 MiB  & 2378 MiB  & 4812 MiB  & 300 MiB \\
    HCSVAE - No Adv.  & 254 MiB  & 255 MiB  & 2558 MiB  & 4588 MiB  & 292 MiB \\
    HCSVAE  & 253 MiB  & 256 MiB  & 2998 MiB  & 4466 MiB  & 294 MiB \\
    DIVA  & 253 MiB  & 255 MiB  & 2634 MiB  & 4996 MiB  & 300 MiB \\
    CCVAE  & 253 MiB  & 255 MiB  & 3066 MiB  & 4998 MiB  & 300 MiB \\
    \methodname\ (ours)  & 254 MiB  & 257 MiB  & 3612 MiB  & 7078 MiB  & 308 MiB \\
    \bottomrule
  \end{tabular}
\end{table}

\subsubsection{Parametric model}
For the parametric model, we consider $z, w \in \mathbb{R}$ and use multi-layer perceptrons (MLPs) with $n_{hidden}=2, d_{hidden} = 8$ to parameterize approximate posteriors, the generative model and classifiers. For all models, we use learning rate $\gamma = 0.001$. A more detailed table of model-specific loss weights is provided in Supplementary Table \ref{suptable:1d_weights}.

\begin{table}[H]
  \caption{Loss weights for the parametric model experiment.}
  \label{suptable:1d_weights}
  \centering
  \begin{tabular}{lcccccc}
    \\ \toprule
    & Rec. & $\KL(Z)$ & $\KL(W)$ & Adv. & Class. & Rec. - ($Z$) \\
    \midrule
    CSVAE - No Adv. & 1 & 1 & 1 & --- & --- & ---\\
    CSVAE & 2.5 & 1 & 0.5 & 20 & --- & --- \\
    HCSVAE - No Adv. & 1 & 1 & 0.5 & --- & --- & --- \\ 
    HCSVAE & 2.5 & 1 & 0.5 & 20 & --- & --- \\
    DIVA & 1 & 1 & 1 & --- & 1 & --- \\
    CCVAE & 1 & 1 & 1 & --- & 1 & --- \\
    \methodname\ (ours) & 0.75 & 0.9 & 0.2 & 0.8 & --- & 0.25 \\
    \bottomrule
  \end{tabular}
\end{table}

\begin{table}[H]
    \caption{K-Means NMI for embeddings across stimulation ($y$) and cell type (common structure). \\ }
  \label{suptable:kangclust}
  \small 
  \centering
  \begin{tabular}{lccc}
    \\ \toprule
    & $w$ - Stimulation ($\uparrow$) & $z$ - Cell Type ($\uparrow$) & $z$ - Stimulation ($\downarrow$) \\
    \midrule
    CSVAE - No Adv. & $\mathbf{0.949 \pm 0.003} $ & $ 0.702 \pm 0.015 $ & $ 0.187 \pm 0.0 $ \\
    CSVAE & $ 0.939 \pm 0.002 $ & $ 0.406 \pm 0.001 $ & $ \mathbf{0.002 \pm 0.0} $ \\
    HCSVAE - No Adv. & $ 0.933 \pm 0.006 $ & $ 0.628 \pm 0.016 $ & $ 0.091 \pm 0.002 $ \\
    HCSVAE & $ 0.931 \pm 0.005 $ & $ 0.433 \pm 0.001 $ & $ 0.003 \pm 0.0 $ \\
    DIVA & $ 0.801 \pm 0.0 $ & $ 0.628 \pm 0.011 $ & $ 0.056 \pm 0.0 $ \\
    CCVAE & $ 0.604 \pm 0.0 $ & $ 0.683 \pm 0.016 $ & $ 0.103 \pm 0.0 $ \\
    \midrule
    \methodname\ (ours) & $ 0.906 \pm 0.003 $ & $ \mathbf{0.716 \pm 0.031} $ & $ \mathbf{0.002 \pm 0.001} $ \\ 
    \bottomrule
  \end{tabular}
\end{table}

\subsubsection{Noisy Swiss roll}
For this experiment, we consider $z, w \in \mathbb{R}^2$ and use MLPs with $n_{hidden}=2, d_{hidden} = 128$ to parameterize approximate posteriors, the generative model and classifiers. For all models, we use learning rate $\gamma = 0.001$. A more detailed table of model-specific hyperparameters is provided in Supplementary Table \ref{suptable:nsr_weights}.

\begin{table}[H]
  \caption{Loss weights for the noisy Swiss roll experiment.}
  \label{suptable:nsr_weights}
  \centering
  \begin{tabular}{lcccccc}
    \\ \toprule
    & Rec. & $\KL(Z)$ & $\KL(W)$ & Adv. & Class. & Rec. - ($Z$) \\
    \midrule
    CSVAE - No Adv. & 20 & 0.2 & 1 & --- & --- & ---\\
    CSVAE & 20 & 0.2 & 1 & 50 & --- & --- \\
    HCSVAE - No Adv. & 20 & 0.2 & 1 & --- & --- & --- \\ 
    HCSVAE & 20 & 0.5 & 1 & 50 & --- & --- \\
    DIVA & 20 & 0.2 & 0.2 & --- & 1 & --- \\
    CCVAE & 20 & 0.2 & 0.2 & --- & 1 & --- \\
    \methodname\ (ours) & 0.9 & 0.2 & 0.2 & 8 & --- & 0.1 \\
    \bottomrule
  \end{tabular}
\end{table}

\subsubsection{Noisy colored MNIST}
For this experiment, we consider $z \in \mathbb{R}^{20}$, $w \in \mathbb{R}^{2}$ and use convolutional neural networks (CNNs) to parameterize approximate posteriors and the generative model. For this example, \methodname{} can support $z, w$ with different sizes, by parameterizing $p(w \mid y)$ through neural networks. For all models, we use learning rate $\gamma = 0.0001$. We detail the architectures and model-specific hyperparameters in Supplementary Tables \ref{suptable:cmnist_enc} - \ref{suptable:noisy_cmnist_weights}. All other neural networks are formulated as MLPs with $n_{hidden}=2, d_{hidden} = 4096$.
\begin{table}[H]
  \caption{Image encoder architecture for noisy colored MNIST. Parameters for Conv2d are input / output channels. Parameters for MaxPool2D are kernel size and stride. Parameter for the linear layer is the output size. For variances, outputs are passed through an additional Softplus layer to ensure non-negativity.}
  \label{suptable:cmnist_enc}
  \centering
  \begin{tabular}{lc}
    \\ \toprule
    Block & Details  \\
    \midrule
    1 & Conv2d(3,32) + BatchNorm2D + ReLU \\
    2 & Conv2d(32,32) + BatchNorm2D + ReLU + MaxPool2D(2,2) \\
    3 & Conv2d(32,64) + BatchNorm2D + ReLU + MaxPool2D(2,2) \\
    4 & Conv2d(64,128) + BatchNorm2D + ReLU + MaxPool2D(2,2) \\
    5 & Linear(4096) + BatchNorm1D + ReLU  \\
    6 & Linear(4096) + BatchNorm1D + ReLU  \\
    7 & Linear($d_{latent}$) \\
    \bottomrule
  \end{tabular}
\end{table}

\begin{table}[H]
  \caption{Image decoder architecture for noisy colored MNIST. Parameters for Conv2d are input / output channels. Parameters for MaxPool2D are kernel size and stride. Parameter for the linear layer is the output size.}
  \label{suptable:cmnist_dec}
  \centering
  \begin{tabular}{lc}
    \\ \toprule
    Block & Details  \\
    \midrule
    1 & Linear(4096) + BatchNorm1D + ReLU  \\
    2 & Linear(4096) + BatchNorm1D + ReLU  \\
    3 & Linear(1152) + Unflatten \\
    4 & Upsample(2) + Conv2d(128, 64) + BatchNorm2D + ReLU \\
    5 & Upsample(2) + Conv2d(64, 32) + BatchNorm2D + ReLU \\
    6 & Upsample(2) + Conv2d(32, 32) + BatchNorm2D + ReLU \\
    7 & Conv2d(32, 3) + Sigmoid \\ 
    \bottomrule
  \end{tabular}
\end{table}

\begin{table}[H]
  \caption{Latent classifier architecture for noisy colored MNIST. Outputs parameterize logits of class probabilities.}
  \label{suptable:cmnist_lat_class}
  \centering
  \begin{tabular}{lc}
    \\ \toprule
    Block & Details  \\
    \midrule
    1 & Linear(4096) + BatchNorm1D + ReLU  \\
    2 & Linear(4096) + BatchNorm1D + ReLU  \\
    3 & Linear(2) \\ 
    \bottomrule
  \end{tabular}
\end{table}

\begin{table}[H]
  \caption{Loss weights for the noisy colored MNIST experiment.}
  \label{suptable:noisy_cmnist_weights}
  \centering
  \begin{tabular}{lcccccc}
    \\ \toprule
    & Rec. & $\KL(Z)$ & $\KL(W)$ & Adv. & Class. & Rec. - ($Z$) \\
    \midrule
    CSVAE - No Adv. & 1 & 0.0001 & 1 & --- & --- & ---\\
    CSVAE & 1 & 0.0001 & 1 & 1 & --- & --- \\
    HCSVAE - No Adv. & 1000 & 0.0001 & 1 & --- & --- & --- \\ 
    HCSVAE & 10000 & 0.0001 & 1 & 1 & --- & --- \\
    DIVA & 1 & 0.0001 & 0.0001 & --- & 1 & --- \\
    CCVAE & 1 & 0.0001 & 0.0001 & --- & 1 & --- \\
    \methodname\ (ours) & 0.5 & 0.0001 & 0.0001 & 0.1 & --- & 0.5 \\
    \bottomrule
  \end{tabular}
\end{table}

\subsubsection{CelebA-glasses}
Motivated by the previous application of \citet{CSVAE}, our choices follow those outlined in \citet{pmlr-v48-larsen16}. We provide a detailed table of model-specific hyperparameters in Supplementary Table \ref{suptable:celeba_weights}:

\begin{table}[H]
  \caption{Loss weights for the CelebA-Glasses experiment.}
  \label{suptable:celeba_weights}
  \centering
  \begin{tabular}{lcccccc}
    \\ \toprule
    & Rec. & $\KL(Z)$ & $\KL(W)$ & Adv. & Class. & Rec. - ($Z$) \\
    \midrule
    CSVAE - No Adv. & 1 & 0.0001 & 1 & --- & --- & ---\\
    CSVAE & 1000 & 0.0001 & 1 & 1 & --- & --- \\
    HCSVAE - No Adv. & 1000 & 0.0001 & 1 & --- & --- & --- \\ 
    HCSVAE & 10000 & 0.0001 & 1 & 1 & --- & --- \\
    DIVA & 100000 & 0.0001 & 0.0001 & --- & 1 & --- \\
    CCVAE & 100000 & 0.0001 & 0.0001 & --- & 1 & --- \\
    \methodname\ (ours) & 1000000 & 0.0001 & 0.0001 & 2000 & --- & 100000 \\
    \bottomrule
  \end{tabular}
\end{table}

\subsubsection{scRNA-Seq}
Following on the previous applications by \citet{scVI}, we use $z\in \mathbb{R}^{10}$, $w \in \mathbb{R}^2$. Similar to the Noisy Colored MNIST example, we use \methodname{} with matched sizes by parameterizing $p(w \mid y)$ through a neural network. We use MLPs with $n_{hidden}=1, d_{hidden}=128$ to parameterize approximate posteriors, the generative model and classifiers. We calculate K-Means NMI through \mbox{\emph{scikit-learn}} \citep{scikit-learn} by calling the \mbox{\texttt{normalized\_mutual\_info\_score}} function with the original labels and the clusterings obtained by running KMeans on (1) the entire latent embedding and (2) single dimensions of the embedding and report the highest score. A more detailed table of model-specific hyperparameters is provided in Supplementary Table \ref{suptable:kang_weights}:

\begin{table}[H]
  \caption{Loss weights for the scRNA-seq experiment.}
  \label{suptable:kang_weights}
  \centering
  \begin{tabular}{lcccccc}
    \\ \toprule
    & Rec. & $\KL(Z)$ & $\KL(W)$ & Adv. & Class. & Rec. - ($Z$) \\
    \midrule
    CSVAE - No Adv. & 1 & 0.0001 & 1 & --- & --- & ---\\
    CSVAE & 1 & 0.0001 & 1 & 100 & --- & --- \\
    HCSVAE - No Adv. & 1 & 0.0001 & 1 & --- & --- & --- \\ 
    HCSVAE & 1 & 0.0001 & 1 & 100 & --- & --- \\
    DIVA & 1 & 0.0001 &  0.0001 & --- & 1 & --- \\
    CCVAE & 1 & 0.0001 & 0.0001 & --- & 1 & --- \\
    \methodname\ (ours) & 0.9 & 0.0001 & 0.0001 & 100 & --- & 0.1 \\
    \bottomrule
  \end{tabular}
\end{table}

\subsection{Summary of the scVI generative model for \ref{res:kang}} \label{sup:scvi}
Given batch key $b$ and $G$ genes, the generative model of scVI for a single cell $x_i \in \mathbb{N}^G$ is formulated as:
\begin{align*}
     & z_i \sim \mathcal{N}(0,1) \\
     & \rho_i = f_{\theta}(z_i, b_i) \\
     & \pi_{ig} = h_{\phi}^g(z_i, b_i) \\
     & x_{ig} \sim \text{ZINB}(l_i\rho_i, \theta_g, \pi_{ig})
\end{align*}
Here, $g$ indexes genes, $l_i = \sum_gx_{ig}$ denotes the total number of counts for a single cell, $z_i$ denotes the latent representation of the cell, and $\rho_i$ denotes the normalized expression of the cell. $f_\theta$ is formulated as a neural network with a final softmax layer. $h_\phi$ is a neural network used to parameterize zero-inflation probabilities for the generative zero-inflated negative binomial (ZINB) distribution. As such, for a single batch, the formulation of scVI is equivalent to the VAE with a ZINB likelihood. While all other models can be extended easily, \methodname\ requires reconstructions as a proxy for the adversarial loss. For this formulation, we directly treat the normalized expressions $\rho_i$ as the adversarial reconstructions $\hat{x}$.

\end{document}